\documentclass[sigconf, screen]{acmart}

\renewcommand\footnotetextcopyrightpermission[1]{}
\AtBeginDocument{%
  }

\acmConference[e-Energy ’26]{The 17th ACM International Conference on Future and Sustainable Energy Systems (ACM e-Energy 2026)}{June 22--25, 2026}{Banff, Canada}

\acmSubmissionID{98}

\usepackage{pifont}
\usepackage{multirow}
\usepackage{listings}       
\usepackage{xcolor}        
\usepackage{tabularx}
\newcolumntype{Y}{>{\centering\arraybackslash}X}
\usepackage{booktabs}       
\usepackage{placeins}
\usepackage{amsmath}
\usepackage{algorithm}
\usepackage{algorithmicx}
\usepackage{algcompatible}
\usepackage{algcompatible}
\usepackage[dvipsnames]{xcolor}
\DeclareMathOperator*{\argmin}{arg\,min}
\usepackage{siunitx}
\usepackage{xcolor}
\usepackage{subcaption}
\usepackage{graphicx}
\usepackage{caption}

\begin{document}

\title{Phythesis: \underline{Phy}sics-Guided Evolutionary Scene Syn\underline{thesis} for Energy-Efficient Data Center Design via LLMs}


  
  

\author{Minghao Li, Ruihang Wang, Rui Tan, Yonggang Wen}
\affiliation{%
  \institution{Nanyang Technological University}
  \country{Singapore}}
\email{minghao002@e.ntu.edu.sg, ruihang001@e.ntu.edu.sg, tanrui@ntu.edu.sg, ygwen@ntu.edu.sg}

\renewcommand{\shortauthors}{Li et al.}

\begin{abstract}
Data center (DC) infrastructure serves as the backbone to support the escalating demand for computing capacity.
Traditional design methodologies that blend human expertise with specialized simulation tools scale poorly with the increasing system complexity.
Recent studies adopt generative artificial intelligence to design plausible human-centric indoor layouts. However, they do not consider the underlying physics, making them unsuitable for the DC design that sets quantifiable operational objectives and strict physical constraints.
To bridge the gap, we propose {\em Phythesis}, a novel framework that synergizes large language models (LLMs) and physics-guided evolutionary optimization to automate simulation-ready (SimReady) scene synthesis for energy-efficient DC design. Phythesis employs an iterative bi-level optimization architecture, where (i) the {\em LLM-driven optimization} level generates physically plausible three-dimensional layouts and self-criticizes them to refine the scene topology, and (ii) the {\em physics-informed optimization} level identifies the optimal asset parameters and selects the best asset combination.
Experiments on three generation scales show that Phythesis achieves 57.3\% generation success rate increase and 11.5\% power usage effectiveness (PUE) improvement, compared with the vanilla LLM-based solution. 
\end{abstract}



\received[accepted]{9 Dec 2025}

\maketitle

\vspace{-0.6em}
\section{Introduction}
\label{sec:intro}
\begin{figure*}[t]
  \includegraphics[width=\linewidth]{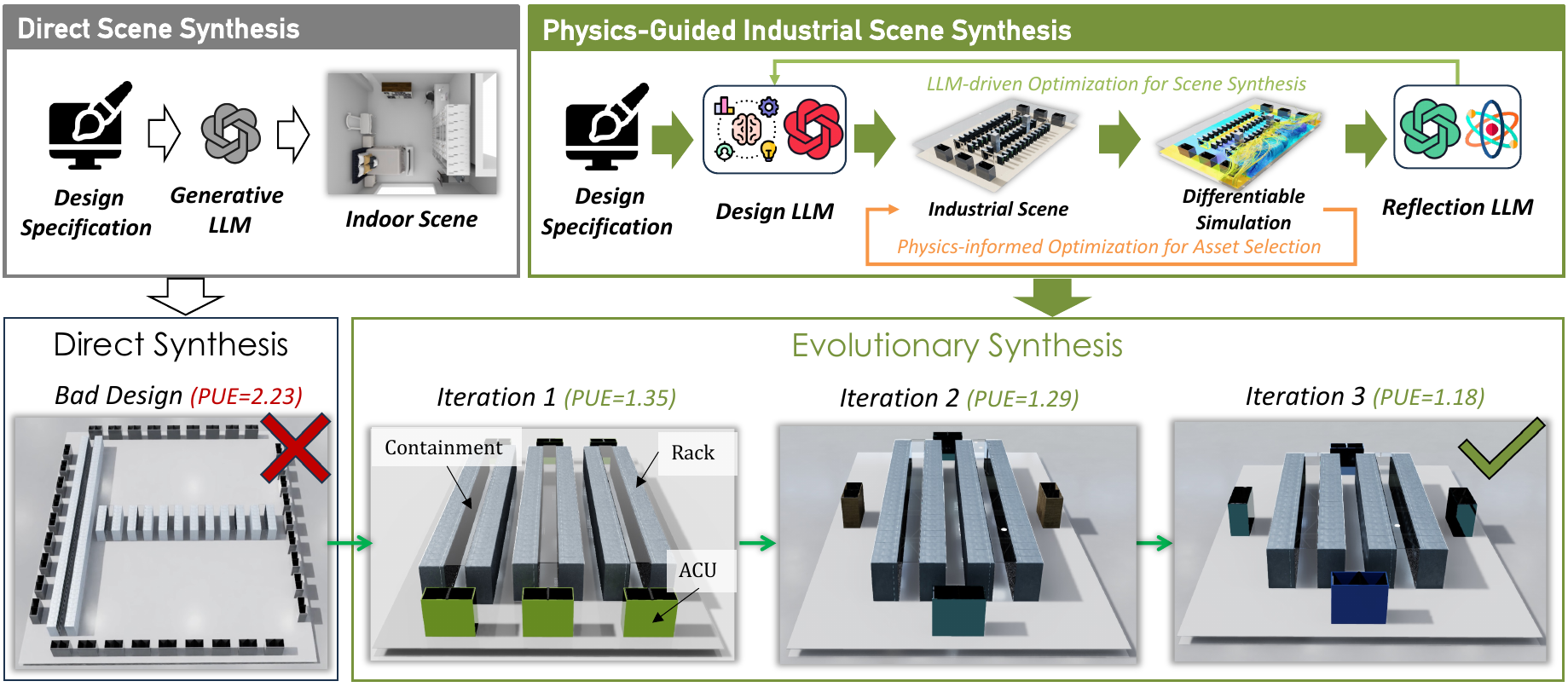}
  \vspace{-2.2em}
  \caption{Our proposed physics-guided scene synthesis framework automates the laborious phases of DC prototype design while enhancing LLMs with physics prior knowledge and physics-guided feedback. We employ a bi-level optimization strategy to iteratively improve LLM-driven generative design, including the level of {\em LLM-driven optimization} for scene synthesis and the level of {\em physics-informed optimization} for asset selection. Note that PUE, i.e., the metric to be optimized, is the ratio of the total energy used by a DC to the energy delivered to its IT equipment. Lower PUE suggests higher energy efficiency.}
  \vspace{-1.4em}
  \label{fig:teaser}
\end{figure*}

The rapid growth of artificial intelligence (AI) applications is driving the increased demand for computing, with a projected annual growth rate of up to 22\% by 2030~\cite{McKinsey2024}.
To meet this growth, data centers (DCs) are being deployed worldwide as the mission-critical infrastructures for the development of the digital economy.
It is predicted that the global demand for new DC constructions will be trippled in the next five years \cite{McKinsey2024}.
This increase presents the need of accelerating design cycles \cite{forbestechcouncil2024}.
However, DC design is a time-consuming process that involves information gathering, design modeling, iterative simulations, and optimizations.
The desired acceleration must ensure that short-term gains in design time do not compromise long-term objectives throughout DC operations, including system reliability and energy efficiency~\cite{shehabi2011data, fakhim2011cooling}. In particular, the energy efficiency, largely dependent on the design, takes effect throughout the entire operation phase that may last for years. It is reported that well-designed DCs can be 30\% more energy-efficient than the average \cite{cai2024towards}.
In sum, the design goals present a multi-objective optimization problem that requires interdisciplinary knowledge and substantial efforts for its solutions.

Existing solutions relevant to DC design can be categorized into traditional semi-automated design and AI-based automated scene synthesis.
Traditional DC design relies on human expertise and specialized tools. 
For example, computational physics-based models are often used to validate the layout, configure equipment, and optimize the thermal management of a DC. 
However, physical system modeling is a tedious process, especially for large-scale DC systems. Manually configuring the geometry or physical properties for an industry-grade data hall that hosts thousands of asset items can take weeks or even months to achieve a satisfactory design~\cite{actiflow2025cfd}.
The AI-driven three-dimensional (3D) scene synthesis approaches shed light on full design automation.
These approaches synthesize indoor 3D scenes by integrating the relationship of objects with geometric and semantic attributes~\cite{qi2018human,gao2024graphdreamer, zhang2024scenepalette}.
In particular, recent studies have investigated using large language models (LLMs) to produce human-competitive designs through iterative prompt engineering and code generation~\cite{feng2023layoutgpt, de2024llmr, du2024text2bim, xia2024scenegenagent}. 
However, these direct scene synthesis methods, as illustrated in Figure~\ref{fig:teaser}, only focus on generating plausible layouts for indoor, warehouse, and building scenarios and do not explicitly consider the underlying physics. This lack of physics considerations renders them unsuitable for DC designs that need to achieve certain objectives defined based on physical variables.
When design objectives and constraints related to physical variables, e.g., temperature and power usage effectiveness (PUE), are specified, the LLMs often produce hallucinations.

To fully implement LLMs' design automation potential and avoid hallucinations, we propose a physics-guided scene synthesis framework called {\em Phythesis} that integrates LLMs with physics-based models for DC design.
At its core, Phythesis uses an evolutionary-style selection loop within a bi-level optimization framework to generate and refine DC design candidates in an iterative manner and eventually yield a final design. 
This approach relies on LLMs' in-context reasoning for exploration rather than genetic operators like crossover and mutation~\cite{forrest1996genetic}.
The bi-level optimization, as illustrated in Figure~\ref{fig:teaser}, consists of (i) {\em LLM-driven optimization} level for scene synthesis and (ii) {\em physics-informed optimization} level for asset selection. The LLM-driven optimization uses two agents, i.e., {\em Design LLM} for scene synthesis and {\em Reflection LLM} for scene refinement. The Design LLM integrates physics priors and design requirements to generate semantic design candidates and output simulation-ready (SimReady)~\cite{nvidia_simready_assets} scenes with 3D geometries and physics attributes. A physics engine performs simulations for these SimReady scenes. The Reflection LLM analyzes the simulation trajectories and critically selects the top candidates in terms of the degree of compliance with the design requirements. The selected designs are used as the context of the Design LLM in the next iteration.
The physics-informed optimization is applied after the Design LLM but before the Reflection LLM to optimize various learnable physical parameters and asset selection based on the simulation results. This optimization is performed efficiently via backpropagation over a differentiable physics-based model.

We evaluate Phythesis through experiments on multiple DC design scenarios against baselines and ablation settings. 
Compared with the vanilla LLM-based direct scene synthesis, Phythesis improves the generation success rate (i.e., the percentage of achieving the physics-related design objectives) by 57.3\% and improves PUE by 11.5\% for successful generations.
Ablation studies reveal that disabling physics-informed optimization or LLM-driven optimization causes the generation success rate to drop by 24\% and expected energy usage to increase by 7\% on three scales.
This validates the necessity of the bi-level optimization design of Phythesis.

Our contributions are summarized as follows:
\begin{itemize}
    \item We propose Phythesis, the first framework that bridges physics guidance with LLMs for DC generative design. 
    Our work resonates with the momentum of generative AI for the design of sustainable energy systems.
    \item We propose a bi-level optimization in the loop of evolutionary scene synthesis, where LLM-driven optimization recommends and refines the scene topology candidates, and physics-informed optimization improves asset selection and parameter configuration. 
    \item We present experiments on multiple design tasks with varied design prompts against several baselines and ablation settings. 
    The results show that combining LLM-driven and physics-informed optimizations effectively improves generation efficiency.
    A case study highlights the potential viability of Phythesis in real-world DC design.
\end{itemize}

{\em Paper organization:} \textsection\ref{sec:related} reviews related work. \textsection\ref{sec:pre} presents preliminary and problem definition. \textsection\ref{sec:design} presents the design of Phythesis. \textsection\ref{sec:eval} and \textsection\ref{sec:case} present evaluation results and a case study, respectively. \textsection\ref{sec:discuss} discusses several related issues. \textsection\ref{sec:conclude} concludes this paper.

\vspace{-0.7em}
\section{Related Work}
\label{sec:related}

\subsection{DC Design \& Scene Synthesis}
Existing studies on DC design cover building construction design and cooling facility selection. 
Fakhim et al.~\cite{fakhim2011cooling} provide insights on cooling facility selection with experimentation in an operational DC.
Shehabi et al.~\cite{shehabi2011data} characterize the differences in local climate and mechanical equipment among DCs and evaluate the impacts of the design on energy usage. 
Recent industry experiences offer new insights into modern DC design and construction. Fakhim et al.~\cite{uptime_tier_topology} outline criteria for differentiating a 4-tier DC site infrastructure topology, requiring comprehensive performance tests for each DC subsystem.
These studies present massive human efforts to design a DC system.

To reduce human effort, automated design methods use generative 3D scenes to arrange objects with real-world spatial relationships and functions. 
Qi et al.~\cite{qi2018human} introduce a human-centric approach for the sampling and synthesis of 3D room layouts. Graph-based methodologies for scene synthesis~\cite{gao2024graphdreamer, zhang2024scenepalette} enable the construction of compositional 3D environments through the utilization of scene graphs.
Recent studies using LLMs to generate 3D scenes introduce various techniques to control scene components.
LayoutGPT~\cite{feng2023layoutgpt}, 3D-GPT~\cite{sun20233d}, and ChatTwin~\cite{li2023chattwin} employ LLMs to directly generate precise geometric configurations, which can be used to render virtual objects and scenes.
Studies like LLMR~\cite{de2024llmr}. SceneGenAgent~\cite{xia2024scenegenagent}, and Text2BIM~\cite{du2024text2bim} use LLM agents to collaborate and reason in the task of transforming textual user input into imperative code for automating 3D scene generation.
While these existing studies demonstrate the potential of LLMs for plausible 3D scene generation, it is still challenging to apply them to DC design that presents physics-related objectives and constraints. The primary reason is that these existing approaches fall short of capturing the physical processes of DC, such as the thermodynamics in data halls. To address this gap, this work combines LLM-driven scene synthesis and physics-informed optimization, such that the generated designs conform to the physics-related design specifications.

\subsection{Physical Model Generation}

While this work studies using LLMs to generate DC designs, the recent studies on using LLMs to generate physical models are relevant.
Early work uses LLMs to automate the configuration of a physical model, which can be in the form of a simulator and neural surrogate, to replicate the real system's attributes.
EPlus-LLM~\cite{jiang2024eplus} fine-tunes LLMs to generate the configuration of an EnergyPlus simulation model from natural language building descriptions, such that the simulation results are close to the measurement traces collected from the real system.
To augment adaptability to undefined scenarios, recent research employs LLMs to generate physics-informed neural networks as models of the real systems.
Ma et al.~\cite{ma2024llm} demonstrate the use of LLMs to explicitly discover governing physics laws from experimental data, thereby enabling interpretable predictions for materials science. 
HDTwinGen~\cite{holt2024automatically} creates hybrid digital twins for dynamic systems by fitting to operational data through an LLM-based approach that generates symbolic representations. 
The primary goal of the studies reviewed above is to generate a physical model that well aligns with the data collected from the real system. Differently, generative DC design aims at generating designs from specifications before the real system is built.

\begin{figure}[t]
    \centering
    \includegraphics[width=\linewidth]{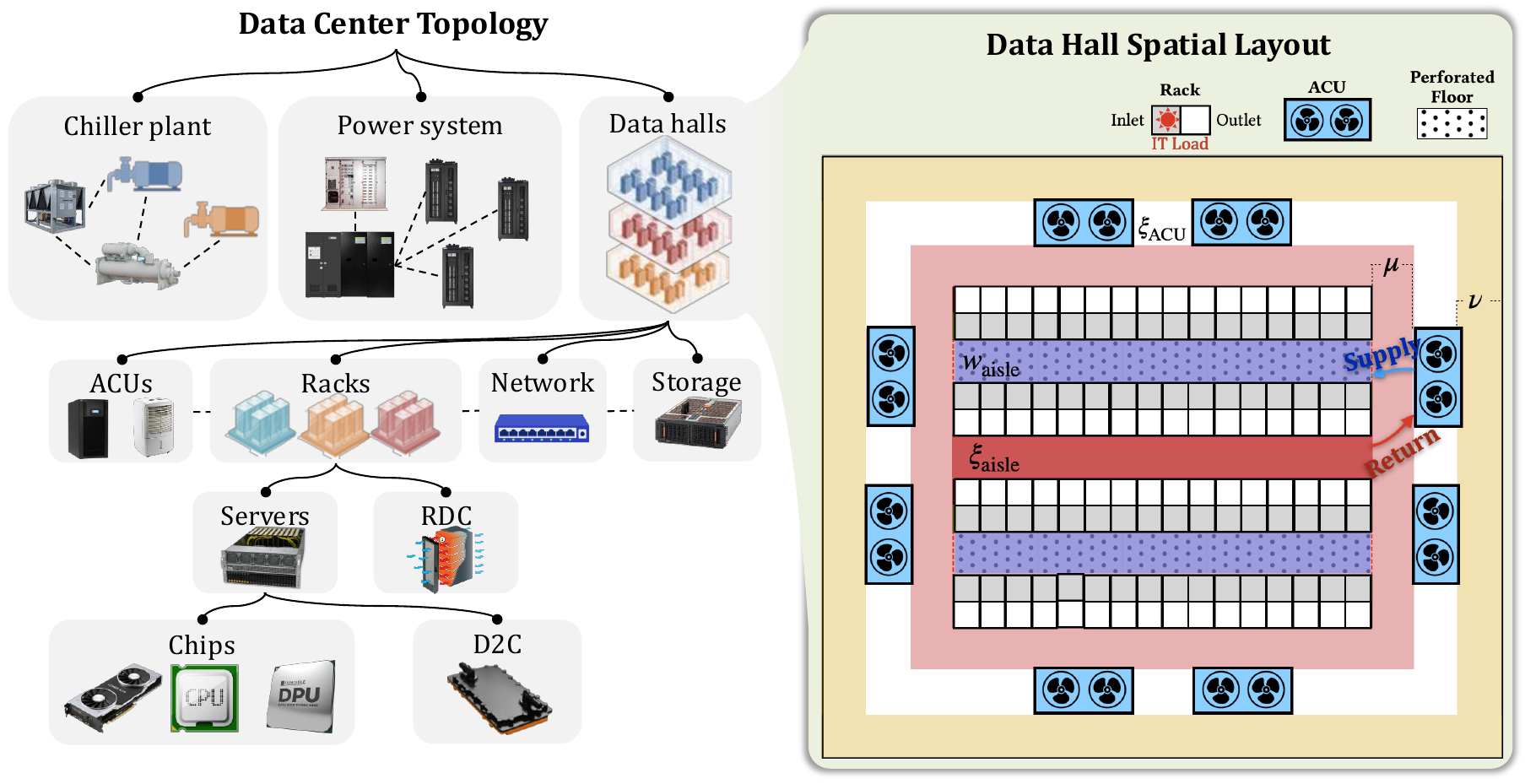}
    \vspace{-2em}
    \caption{A typical DC topology integrates cooling and power distribution systems to support computing services held by ITE in a data hall with a specific layout.}
    \label{fig:dc_struct}
\end{figure}

\section{Preliminaries and Problem Definition}
\label{sec:pre}
In this section, we introduce the preliminaries including DC architecture and digital twin systems. Then, we define the problem of DC generative design.

\subsection{Preliminaries}
\subsubsection{DC system architecture}
As illustrated in Figure~\ref{fig:dc_struct}, the typical DC topology is structured around three core systems, i.e., the chiller plant, power distribution system, and data halls. The chiller plant is responsible for removing the heat generated from the data hall that hosts information technology equipment (ITE). The power distribution system delivers stable electricity via uninterruptible power supply (UPS) units and power distribution cabinets to all active components. Within the data halls, the ITE, including servers, network equipment, and storage systems, is organized into racks. Servers are equipped with advanced computing chips to provide computing services. The heat generated by servers can be removed via indoor air cooling or direct-to-chip (D2C) liquid cooling systems.

Compared with synthesizing human-centric indoor environments and outdoor landscapes, industrial scene synthesis is a more complex task.  Designing a DC requires consideration of multiple factors, including site selection, equipment configuration, governance regulations, and environmental conditions. The integrative consideration of these factors is important to the reliability, sustainability, and cost-effectiveness of the DC once it becomes operational.
In this paper, we focus on the 3D scene synthesis of the data hall and the selection of associated cooling equipment based on design specifications. The design automation aims to maximize energy efficiency through optimal equipment combinations while adhering to data hall layout principles and other operational constraints.

\subsubsection{DC digital twin system}
\begin{figure}[t]
    \centering
    \includegraphics[width=\linewidth]{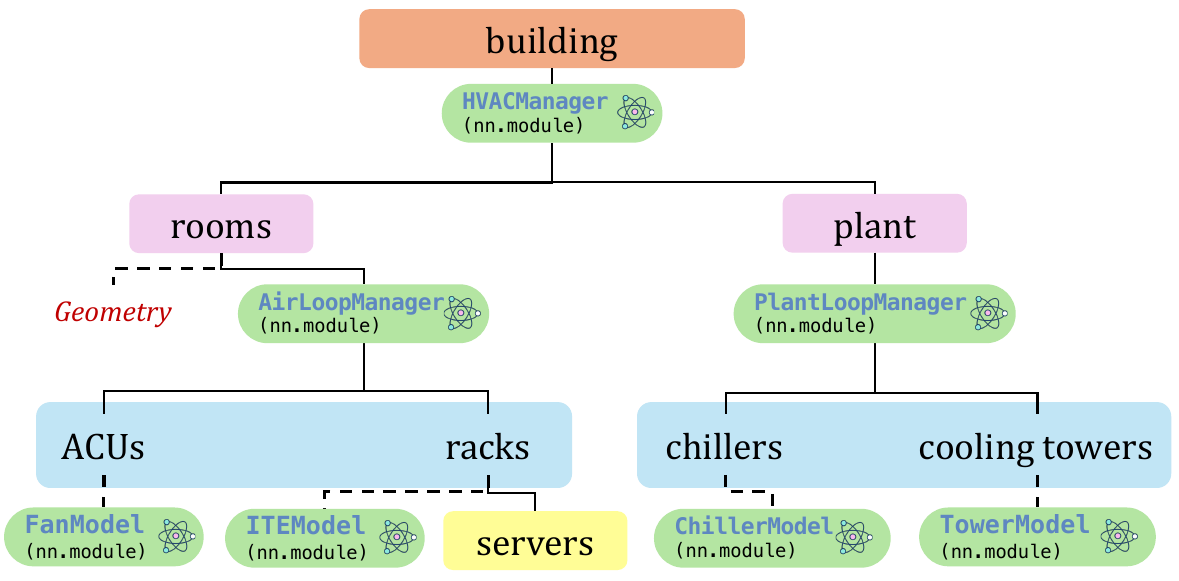}
    \caption{The hierarchical structure of the DC digital twin. A digital twin consists of a scene description and physics-based models for each SimReady asset.
    }
    \label{fig:dtfile}
\end{figure}
Digital twins are virtual replicas of physical entities that practitioners use to design, simulate, and operate their physical counterparts~\cite{nvidia_simready_assets}.
In this paper, we adopt the concept of digital twin for system design. It is based on a scene description file specifying the configuration and the corresponding physics-based models for all SimReady assets in the system.
A DC digital twin hierarchically manages subsystem connections and simulations, creating a multi-level framework as shown in Figure~\ref{fig:dtfile}.

The first level represents the entire "\texttt{Building}"; the second level encompasses "\texttt{Rooms}" and "\texttt{Plant}"; the third level includes specific assets within these categories. The Phythesis framework focuses on a subset of assets, including "\texttt{Racks}" and "\texttt{ACUs}" within "\texttt{Rooms}", and "\texttt{Chillers}" and "\texttt{Towers}" within "\texttt{Plant}". An air-cooling unit (ACU) is a precise cooling system that maintains the indoor air temperature for the stable operation of ITE. A cooling tower removes waste heat from water by evaporative cooling. A chiller removes heat from a liquid via a refrigeration cycle, supplying chilled water to the ACUs. This hierarchical scene description ensures a clear and organized representation of the DC's physical components.
To enable accurate physical simulation based on the DC digital twin, each asset is associated with a differentiable physics-based model. 
These models capture the physical behaviors of and interactions among individual assets, including their thermodynamics, power usage profiles, and cooling properties. 
In the upper class of these facilities, "\texttt{AirLoopManager}" and "\texttt{PlantLoopManager}" capture the interactions between each inner asset.
On the top level, "\texttt{HVACManager}" manages the simulation and interactions between "\texttt{Rooms}" and "\texttt{Plant}".
By integrating models into the hierarchical structure, the digital twin facilitates precise and differentiable simulations, enabling optimization and real-time analysis of the physical environment of the DC. We defer the detailed discussion regarding the differentiable property of the digital twin system to \textsection\ref{subsec:phy-opt}.

\subsection{Problem Definition}
This section formalizes the design problem. We denote a DC scene as a tuple of scene topology $\mathcal{T}$, spatial layout $\mathcal{D}$, and asset combinations $\mathcal{A}$:
\begin{equation}
\label{eq:scene}
    S = (\mathcal{T}, \mathcal{D}, \mathcal{A}).
\end{equation}
The scene topology is defined with nodes $V$ and edges $E$ as $\mathcal{T} = (V, E)$, where nodes set $V = \{v_1, \dots, v_n\}$ consists of asset identifiers, edges refer to the connection relationships as illustrated in the left of Figure~\ref{fig:dc_struct}.
The spatial layout of a DC with multiple data halls is denoted as $\mathcal{D}=\{d_1,\dots,d_m\}$, where each $d$ represents the spatial layout of a single data hall.
As illustrated in Figure~\ref{fig:dc_struct}, the spatial layout defines the geometric attributes of a data hall as a tuple $d = (\mu, \nu, w_\text{aisle}, \xi_\text{aisle}, \xi_\text{ACU})$, including the aisle width $w_\mathrm{aisle}$, rack gap $\xi_\mathrm{aisle}$, ACU gaps $\xi_\mathrm{acu}$, margin $\mu$, and padding $\nu$.
The set of selected assets is denoted as $ \mathcal{A} = \{a_1, \dots, a_n\} $, where each asset $a$ consists of its 3D geometric information and physical properties. 
In this paper, we focus on synthesizing the DC scene to optimize a certain performance metric while fulfilling its design requirements.
Without loss of generality, this work primarily uses PUE as the performance metric. It is a widely adopted DC energy efficiency metric, defined as the ratio of total facility energy to ITE energy usage. 
Let $\mathcal{R} \subset \mathcal{A}$ denote racks and $\mathcal{C} \subset \mathcal{A}$ denote cooling units. For servers, let $\mathcal{E}_r$ be servers in rack $r \in \mathcal{R}$, and $\mathcal{E}_h$ be servers in a data hall $h$.
For ACUs, let $\mathcal{C}_h$ be the ACUs in a data hall $h$.
A real-world DC is a complex, heterogeneous infrastructure with diverse server types, power delivery constraints, and multi-tenant requirements. 
To establish a proof of concept for our approach, this paper simplifies the context by assuming homogeneous servers hosted within air-cooled data halls.
We formulate the DC scene synthesis problem as to minimize PUE while meeting three core constraints related to geometry, power, and cooling provision:
\begin{equation}
\label{eq:problem}
\begin{aligned}
& \underset{S}{\min} 
& \quad & \text{PUE}(S), \\
& \text{s.t.} 
& \quad & l_a \in \Omega, \quad \forall a \in \mathcal{A}, \\
& 
& \quad & \sum_{s \in \mathcal{E}_r} p_s \leq p_r, \quad \forall r \in \mathcal{A}, \\
& 
& \quad & \sum_{s \in \mathcal{E}_h} h_s \leq \sum_{c \in \mathcal{C}_h} H_c, \quad \forall c \in \mathcal{A},
\end{aligned}
\end{equation}
where $l_a$ is the location of asset $a$, $\Omega$ is the feasible region of the scene $S$ calculated by $\mathcal{D}$ as illustrated in Figure~\ref{fig:dc_struct}, $p_s$ is the rated power of server $s$, $p_r$ is the power capacity of rack $r$, $h_s$ is the heat output of server $s$, $H_c$ is the cooling capacity of ACU $c$. 

\section{Phythesis Design}
\label{sec:design}
\begin{figure*}[t]
    \centering
    \includegraphics[width=\linewidth]{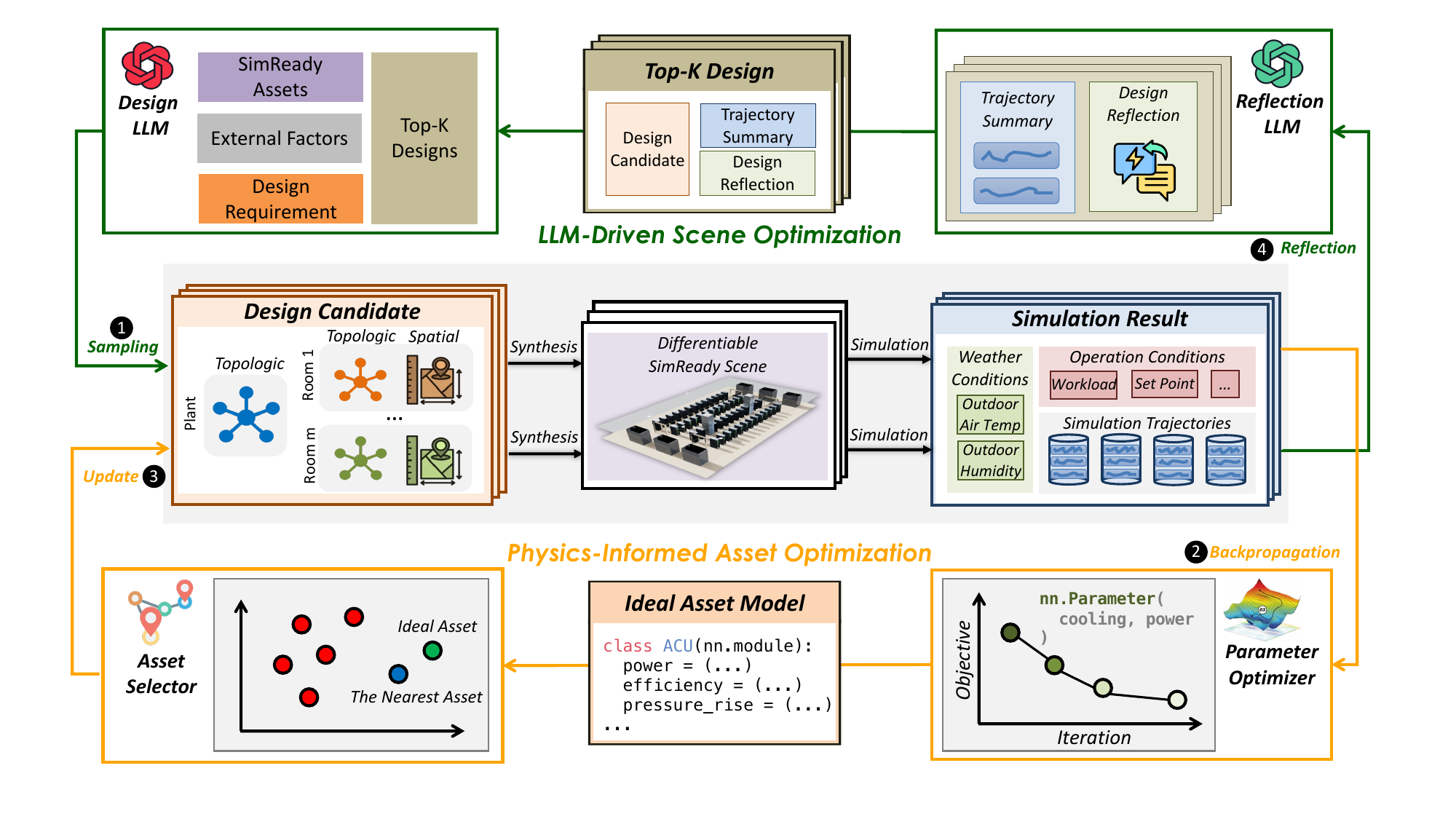}
    \vspace{-2em}
    \caption{Phythesis's framework for scene synthesis uses a physics-guided, bi-level optimization approach. The Design LLM creates semantic design candidates, providing spatial topologies and asset combinations (\ding{202}). These are turned into SimReady scenes with physics attributes for simulation. The physics engine generates trajectories under varying conditions. Phythesis optimizes asset parameters via backpropagation (\ding{203}) and selects optimal assets based on distance (\ding{204}). The top asset combination is re-simulated. The Reflection LLM analyzes time-series data, failures, and metrics (\ding{205}), ranking top-$K$ candidates for refinement. Successful designs are added to the Design LLM's context for further evolutionary synthesis through feedback.}
    \label{fig:framework}
\end{figure*}

In this section, we present the system design of Phythesis, which integrates LLMs with physics-based models for physics-guided evolutionary scene synthesis. 
As illustrated in Figure~\ref{fig:framework}, this approach employs a bi-level optimization framework. 
On the level of {\em LLM-driven optimization}, the Design LLM generates DC scenes by reasoning with physics priors and user prompts; the Reflection LLM summarizes the simulation trajectories for evolutionary synthesis. 
On the level of {\em physics-informed optimization}, physics-based models validate and refine these designs through differentiable simulation and asset optimization. 
The synergy between the two levels of optimization enables Phythesis to balance innovative design exploration and rigorous physics-related constraint adherence.
In this section, we present the following workflow steps:
LLM-driven generative design (\textsection\ref{subsec:gen-design}), physics-informed asset optimization (\textsection\ref{subsec:phy-opt}), and LLM-driven scene optimization (\textsection\ref{subsec:sce-opt}).

\begin{figure}[t]
    \centering
    \includegraphics[width=\linewidth]{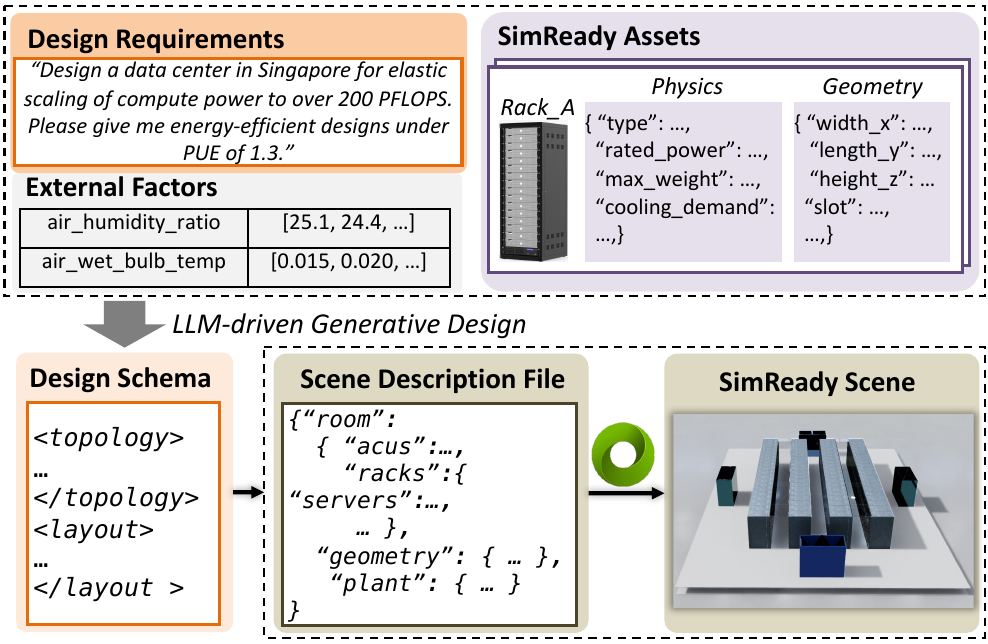}
    \vspace{-2em}
    \caption{LLM-driven generative design (\ding{202}). The Design LLM generates semantic design candidates using natural language design requirements, external factors, and SimReady assets. The design candidates are formatted in a certain schema, which converts to description files for 3D rendering.}
    \label{fig:input_data}
\end{figure}

\subsection{LLM-Driven Generative Design}
\label{subsec:gen-design}
To comprehend user design prompts and translate them into DC scenes, we introduce the Design LLM that incorporates in-context prior knowledge of DC physics.
Our emphasis is on enabling the LLM to generate design candidates by reasoning with physics priors.
First, the prompt schema is formed using design requirements, external factors, and contextual assets,
as shown in Appendix~\ref{sec:prompt}. 
The Design LLM is required to generate a batch of design samples (\ding{202}). 
The overall input and output of the LLM-driven generative design process are shown in Figure~\ref{fig:input_data}. 
Then, this response semantically describes the scene topology, spatial relations, and asset combinations in a structured data format.
Finally, Phythesis converts an abstract design blueprint into a SimReady digital scene by placing 3D models with physics attributes.

\subsubsection{Physics-prior contextualization} 
LLMs have demonstrated remarkable reasoning capabilities in understanding the physical world with contextualized knowledge~\cite{xu2024penetrative}.
To enable robust comprehension of the physical environments in DC, Phythesis adopts in-context learning to contextualize physics priors.
\textit{SimReady Assets}, denoted as $P_\text{sa}$, serve as the foundational building blocks, encapsulating geometric attributes and physical properties~\cite{nvidia_simready_assets}.
\textit{External Factor Profiles}, denoted as $P_\text{fp}$, encompass environmental conditions that affect operational performance.
Factors like the regional weather conditions define localized atmospheric conditions critical for physics-aware design optimization, including air relative humidity and air wet-bulb temperature. 
These parameters directly govern facility selection and construction design strategies.
\textit{User-defined Design Requirements}, denoted as $P_\text{dr}$, include the design objectives and constraints of a DC. It often includes descriptions of computational performance thresholds and efficiency constraints. 
We define a certain prompt template $C_\text{design}$ for contextualization,
as shown in Appendix~\ref{sec:prompt}.
To enhance design generation, query $Q_\text{design}$ can be optionally augmented with the top-$K$ historical solutions from the evolutionary heap $H$, i.e.,
$
Q_\text{design} = C_\text{design}(P_\text{sa},P_\text{fp},P_\text{dr},H.\text{topk}(K))
$.

\subsubsection{Design candidate sampling}
The Design LLM generates preliminary design candidates by integrating physics-prior contextualization. 
We employ the nucleus sampling policy to generate a batch of design candidates~\cite{liu2023reason}.
We represent this sampling process by:
\begin{equation}
\{(\mathcal{T}_i, \mathcal{D}_i, \mathcal{A}_i)\}_{i=1}^N \sim \Theta_\text{design}(Q_\text{design}),
\end{equation} 
where $N$ is the number of candidates and $\Theta_\text{design}$ is the Design LLM.
As shown in Figure~\ref{fig:framework}, each candidate defines a comprehensive DC scene as Equation~(\ref{eq:scene}).
The scene topology defined as $\mathcal{T}$ in Equation~(\ref{eq:scene}) establishes the hierarchical structure and connectivity of components within the DC. The response defines parent-child relationships between systems, ensuring logical grouping and dependency alignment. 
This hierarchical organization extends to the indoor layout of the DC, where the data hall itself forms a tree-like structure as illustrated in Figure~\ref{fig:dc_struct}. 
The spatial relations $\mathcal{D}$ govern the precise placement and spacing of assets within the physical environment. 
The Design LLM estimates the minimum spatial distance for safe heat dissipation with $P_\text{sa}$ and $P_\text{fp}$ that satisfies certain industry standard~\cite{GB50174-2017}.

\subsubsection{Domain-specific scene synthesis} 
The synthesis process integrates scene topology, spatial relations, and asset combinations into a physics-enabled cohesive 3D DC scene.
As DC is an industrial system with strict layout principles, we consider the industry-specific rules to synthesize the DC scene. 
This synthesis first translates the abstract topology to a structured scene that defines only system connections without detailed geometrical placement.
Then, it calculates the feasible region $\Omega$  with the spatial relations $\mathcal{D}$ of all data halls, where each selected asset $a$ will be assigned its geometric location $l_a$. This calculation adheres to rule-based constraints considering thermal efficiency as defined in~\cite{li2023chattwin}.
Each asset is retrieved by the identification described in the set of nodes $V_i$ of each sampled scene topology $\mathcal{T}_i$. 
Finally, the asset combination $\mathcal{A}_i$ for the $i$-th design candidate is assembled to a 3D scene $S_i$ as interconnected units with physical properties, enabling analysis of heat transfer and energy simulation. 
Appendix~\ref{sec:asset} gives an example of a SimReady ACU asset.
Phythesis generates the scene $S_i$ for each sampled design as:
\begin{equation}
    S_i = (\mathcal{T}_i, \mathcal{D}_i, \mathcal{A}_i), \quad \forall i \in \{1, \dots, N\}.
\end{equation}
Each generated scene serves as an executable digital twin capable of performing differentiable simulations using Pytorch. We denote the physics-based model of scene $S$ as $\theta_S$.

\subsection{Physics-Informed Asset Optimization}
\label{subsec:phy-opt}
To ensure adherence to physics constraints and enhance design performance, Phythesis incorporates physics-informed asset optimization. This approach focuses on mitigating the LLM's limited mathematical capabilities by integrating physics-informed methods.
Given that the simulation relies on a differentiable physics engine, we first optimize the asset parameters using backpropagation from the simulation output (\ding{203}), thereby generating an ideal asset with optimized parameters.
Phythesis then computes the distance between these ideal parameters and those of each asset in the library, selecting the closest match (\ding{204}). The library contains assets available to the DC constructor.
Finally, this updated asset combination refines the design candidate, allowing Phythesis to generate new SimReady scenes for LLM-driven optimization.

\subsubsection{Simulation forward}
The physics engine simulates heat transfer and energy usage to generate trajectory data for synthetic DC scenes, as defined in Equation~(\ref{eq:scene}). The simulation operates under two distinct sets of conditions: \textit{external conditions} and \textit{operational conditions}. External conditions $I_\text{fp}$, contextualized as $P_\text{fp}$, encompass environmental factors including outdoor air temperature and relative humidity.  
The \textit{operational conditions} and $I_\text{op}$ regulate energy use and thermal dynamics, covering factors like server utilization rate, ACU fan flow rate, and cooling air setpoint. 
The simulation forward process is formalized as:
\begin{equation}
\label{eq:sim}
    (y, z) = \theta_S(I_\text{fp}, I_\text{op}),
\end{equation}
where $\theta_S$ represents the physics-based model for the entire DC simulation of scene $S$. 
The inputs are external conditions $I_\text{fp}$ and operational conditions $I_\text{op}$. The outputs are the simulated physical phenomena $y$ (e.g., zone air temperature) and auxiliary observational feedback $z$ (e.g., PUE). Phythesis adopts a differentiable DC model $\theta_S$ that was developed in our prior work [citation omitted for anonymity]. Specifically, in this DC model, the thermal outputs and energy usages of ACUs, cooling towers, and chillers are continuous and differentiable with respect to the design parameters of each asset in $\mathcal{A}$. This differentiable property enables the efficient optimization of the design parameters as presented shortly.

\subsubsection{Asset parameter optimization}
\label{sec:asset_param_opt}
This step aims to find the ideal continuous parameters for each asset in the design to minimize PUE.
Appendix~\ref{sec:asset} provides an example of an ACU's continuous asset parameters. 
Formally, we optimize the parameters $\boldsymbol{\alpha} = \{\alpha_1, \alpha_2, \dots, \alpha_n\}$ of each asset $a_i \in \mathcal{A}$ in Equation~(\ref{eq:scene}) by solving:
\begin{equation}
    \boldsymbol{\alpha}^* = \argmin_{{\boldsymbol{\alpha}}} z, \quad \text{s.t.} ~G(y) \leq 0.
\end{equation}
The constraint $G(y) \leq 0$ enforces physics-based boundaries, such as energy conservation and material stability, ensuring compliance with the power and cooling constraints defined in Equation~(\ref{eq:problem}). Optimization is achieved through backpropagation using the differentiable $\theta_S$, producing the set of optimal parameters $\boldsymbol\alpha^*$ that minimizes PUE while satisfying all constraints.
The optimized parameters $\alpha^*_j \in \boldsymbol\alpha^*$ are then integrated into the physics-based models of each SimReady asset $a_j \in \mathcal{A}$ in the DC scene. 
This integration updates the original models to reflect the optimized asset configurations, resulting in a refined physics-based model. The update process is formalized as ${a_j}^* = {a_j}(\alpha_j^*)$, where ${a_j}^*$ denotes the ideal asset model that incorporates the optimized parameters. 
Phythesis gets a set of asset models with ideal parameters denoted as $\mathcal{A}^*=\{{a_1}^*,{a_2}^*,\dots,{a_n}^*\}$.
In sum, this step produces a blueprint of ideal inner component specifications. Since $\mathcal{A}^*$ may contain parameters not available in real-world catalogs, the subsequent step maps this blueprint to available assets in the library.

\subsubsection{Asset selection}
Following optimization of asset parameters, Phythesis identifies the closest asset from the asset library to approximate the ideal asset models $\mathcal{A}^*$. 
For each ideal asset ${a_j}^* \in \mathcal{A}^*$ with optimized parameters $\alpha_j^*$, we compute the Euclidean distance in the parameter space between $\alpha_j^*$ and the parameters of all assets in the corresponding category within the library. Specifically, let $L_j$ denote the subset of the asset library $\mathcal{L}$ containing assets of the same category as ${a_j}^*$, where each library asset $b_k \in L_j$ has the same dimension of parameters $\beta_k$. 
The asset that best matches the ideal asset $a_j^*$, represented by $\alpha_j^*$, is selected by minimizing their Euclidean distance in the parameter space:
\begin{equation}
    \hat{a}_j = \argmin_{b_k \in L_j} \| \alpha_j^* - \beta_k \|_2,
\end{equation}
where $\hat{a}_j$ represents the selected asset to replace $a_j$ in the original asset set $\mathcal{A}$.
This process is repeated for all ideal assets, resulting in an updated asset combination $\hat{\mathcal{A}} = \{\hat{a}_1, \hat{a}_2, \dots, \hat{a}_n\}$. The refined design candidate incorporates $\hat{\mathcal{A}}$ to generate a new SimReady scene for further simulation and optimization.
Phythesis updates the scene $S$ with the newly selected assets $\hat{\mathcal{A}}$, while keeping the topology $\mathcal{T}_i$ and layout $\mathcal{D}_i$, as:
\begin{equation}
    \hat{S}_i = (\mathcal{T}_i, \mathcal{D}_i, \hat{\mathcal{A}}_i), \quad \forall i \in \{1, \dots, N\}.
\end{equation}
The physics-informed asset optimization selects the best asset combination under the scene topology $\mathcal{T}$ and layout $\mathcal{D}$.
With the new scene $\hat{S}$ and its physics-based model $\theta_{\hat{S}}$, Phythesis re-conducts a simulation $\theta_{\hat{S}}(I_\text{fp}, I_\text{op})$ for the next step of the workflow, i.e., LLM-driven scene optimization.

\subsection{LLM-Driven Scene Optimization}
\label{subsec:sce-opt}
To reduce the physical hallucinations of the Design LLM, Phythesis incorporates the Reflection LLM to perceive the simulation trajectories and criticize the design candidates. 
Our focus is on enabling LLMs to understand the physical system using simulated sensing data. First, a trajectory is simulated under the assets' design configurations and external factor profiles through the physics engine. The Reflection LLM is required to generate evaluations and suggestions for each design candidate (\ding{205}). Eventually, Phythesis appends those reflection-design parts to a top-$K$ design candidates for the following iterations.

\subsubsection{Simulation data contextualization}
LLMs have demonstrated their abilities to comprehend the physics world via sensing data~\cite{xu2024penetrative}.
The Reflection LLM receives time-series simulation data, failure events, and overall metrics to diagnose systemic flaws from the simulation output $(y,z)$ in Equation~(\ref{eq:sim}). 
It examines cross-domain causality to deliver natural language critiques and summaries of simulation trajectories for each design candidate, using time-series data, failure events, and overall metrics.
\textit{Time-Series Data}, denoted as $P_\text{ts}$, capture system dynamics over time, such as temperature changes in a thermal zone or power use in a DC electrical system. 
\textit{Failure Events}, denoted as $P_\text{fe}$, represent specific incidents where components exceed operational thresholds, such as server rack hotspots triggering shutdowns.
\textit{Overall Metrics}, denoted as $P_\text{om}$, aggregate system performance into key indicators, such as PUE, for DC energy efficiency. 
We design a prompt template $C_\text{reflect}$ to contextualize the physics feedback for the Reflection LLM, as shown in the Appendix~\ref{sec:prompt}.
The physics feedback query $Q_\text{reflect}$ for the Reflection LLM can be written as $Q_\text{reflect}=C_\text{reflect}(P_\text{ts},P_\text{fe},P_\text{om})$.

\begin{algorithm}[t]
\caption{Phythesis Evolutionary Scene Synthesis Algorithm}
\label{alg:phythesis}
\begin{algorithmic}[1]
\STATE \textbf{Inputs:} SimReady asset library $\mathcal{L}$ and its context $P_\text{sa}$, factor profiles $P_\text{fp}$, design requirements $P_\text{dr}$.
\STATE \textbf{Settings:} total iterations $M$, the number of LLM samples $N$, the number of top elements $K$.
\STATE $H \leftarrow \text{heap}()$
\FOR{iteration $m$ in $M$}
    \Statex \textcolor{ForestGreen}{$\triangledown$ LLM-Driven Generative Design}
    \STATE $Q_{\text{design}} \leftarrow C_{\text{design}}(P_{\text{sa}}, P_{\text{fp}}, P_{\text{dr}}, H.\text{topk}(K))$
    \STATE $\{(\mathcal{T}_i, \mathcal{D}_i, \mathcal{A}_i)\}_{i=1}^N \leftarrow \Theta_{\text{design}}(Q_\text{design})$
    \STATE $\{S_i\}_{i=1}^N \stackrel{\text{synthesis}}{\longleftarrow}\{(\mathcal{T}_i, \mathcal{D}_i, \mathcal{A}_i)\}_{i=1}^N$
    \Statex \textcolor{ForestGreen}{$\triangledown$ Physics-Informed Asset Optimization}
    \FOR{$i$ in $N$}
        \STATE $(y_i, z_i) \leftarrow \theta_{S_i}( I_\text{fp}, I_\text{op})$
        
        \STATE $\boldsymbol\alpha^* \leftarrow \argmin_{\boldsymbol\alpha}z_i, \quad \text{s.t.} ~G(y) \leq 0$
        \STATE $\mathcal{A}_i^* \leftarrow \{a_j(\alpha_j^*) | a_j \in \mathcal{A}_i\}$
        \FOR{$a_j^* \in \mathcal{A}^*_i$}
            \STATE $\hat{a}_j \leftarrow \argmin_{b_k \in L_j} \| \alpha_j^* - \beta_k \|_2$
        \ENDFOR
        \STATE $\hat{\mathcal{A}}_i \leftarrow \{\hat{a}_1, \hat{a}_2, \dots, \hat{a}_n\}$
        \STATE $\hat{S}_i \stackrel{\text{synthesis}}{\longleftarrow} (\mathcal{T}_i, \mathcal{D}_i, \hat{\mathcal{A}}_i)$
        \Statex \textcolor{ForestGreen}{$\triangledown$ LLM-Driven Scene Optimization}
        \STATE $P_{\text{ts},i},P_{\text{fe},i},P_{\text{om},i} \leftarrow \theta_{\hat{S}_i}(I_\text{fp}, I_\text{op})$
        \STATE $Q_{\text{reflect},i} \leftarrow C_\text{reflect}(P_{\text{ts},i},P_{\text{fe},i},P_{\text{om},i})$
        \STATE $R_i \leftarrow \Theta_\text{reflect}(\hat{S}_i, Q_{\text{reflect},i})$
    \ENDFOR
    \STATE $\mathbf{R} \leftarrow \{R_i, \hat{S}_i\}_{i=1}^N$
    ; $H.\text{append}(\textbf{R})$
\ENDFOR
\STATE \textbf{RETURN} $H.\text{topk}(1)$
\end{algorithmic}
\end{algorithm}

\subsubsection{Evolutionary design with physics feedback}
Phythesis prioritizes top-$K$ candidates and appends them into the Design LLM’s context window. 
The Top-$K$ design triplet list refers to candidates ranked in ascending order based on the specified energy criteria.
The Reflection LLM evaluates each design candidate and provides design insights. The reflection outputs are written as:
\begin{equation}
\mathbf{R} =\{R_i,\hat{S}_i\}_{i=1}^N,~R_i \sim \Theta_\text{reflect}(\hat{S}_i,Q_{\text{reflect},i}),
\end{equation} where $R_i$ is the semantic reflection output with trajectory summary and design reflection for the scene $S_i$, and $\Theta_\text{reflect}$ is the Reflection LLM.
Each top-$K$ design history is a triplet of the semantic design candidate, the trajectory summary, and the design reflection. 
A heap $H$ appends history $\mathbf{R}$ at the end of each iteration.
Based on the Reflection LLM's evaluations and suggestions, the Design LLM generates new design candidates with updated scene topology, asset combination and spatial layout.
Each new candidate defines a DC scene as in Equation~(\ref{eq:scene}) with updated topology, layout, and asset combinations. Phythesis's evolutionary scene synthesis algorithm is shown in Algorithm~\ref{alg:phythesis}.

\section{Performance Evaluation}
\label{sec:eval}
This section presents the evaluation of Phythesis through different generation scenarios against baselines and ablation settings. The experiments are conducted under different backbone LLMs. 

\subsection{Experiment Setup}
To evaluate the effectiveness of Phythesis in generating realistic and optimized DC scenes, we conduct experiments across a variety of design prompts and scenarios. These settings test the system's capability to integrate semantic intent with domain physics knowledge, ensuring the quality of the generated designs.

\begin{table*}[t]
\centering
\caption{Performance evaluation of scene synthesis methods. Methods are categorized by generation paradigm (Direct/Iterative) and evaluated across computing scales (Small-Edge/Medium-Cluster/Large-Cloud) with two different backbone models (I-LM/R-LM). Baselines (i)-(ii) and ablation setting (vii) are heuristic methods without using LLMs. The metrics include PUE and GSR. Arrows indicate optimization direction ($\uparrow$=higher better, $\downarrow$=lower better). Best values in \textbf{bold} demonstrate statistical significance.}
\label{tab:performance}
\vspace{-1em}
\begin{tabular}{@{}lc|cccc|cccc|cccc@{}}
\toprule
\multirow{3}{*}{\textbf{Method}} & 
\multirow{3}{*}{\textbf{Type}} & 
\multicolumn{4}{c|}{\textbf{Small-Edge}} & 
\multicolumn{4}{c|}{\textbf{Medium-Cluster}} & 
\multicolumn{4}{c}{\textbf{Large-Cloud}} \\
\cmidrule(lr){3-6} \cmidrule(lr){7-10} \cmidrule(lr){11-14}
& & \multicolumn{2}{c}{PUE($\downarrow$)} & \multicolumn{2}{c|}{GSR($\uparrow$)} & \multicolumn{2}{c}{PUE($\downarrow$)} & \multicolumn{2}{c|}{GSR($\uparrow$)} & \multicolumn{2}{c}{PUE($\downarrow$)} & \multicolumn{2}{c}{GSR($\uparrow$)} \\ 
\cmidrule(lr){3-6} \cmidrule(lr){7-10} \cmidrule(lr){11-14}
& & I-LM & R-LM & I-LM & R-LM & I-LM & R-LM & I-LM & R-LM & I-LM & R-LM & I-LM & R-LM \\ 
\midrule
(i) Random & Direct & \multicolumn{2}{c}{1.800} & \multicolumn{2}{c|}{0.293} & \multicolumn{2}{c}{1.338} & \multicolumn{2}{c|}{0.427} & \multicolumn{2}{c}{1.302} & \multicolumn{2}{c}{0.429} \\
(ii) EA & Iterative & \multicolumn{2}{c}{1.796} & \multicolumn{2}{c|}{0.624} & \multicolumn{2}{c}{1.277} & \multicolumn{2}{c|}{0.655} & \multicolumn{2}{c}{1.246} & \multicolumn{2}{c}{0.630} \\
(iii) Vanilla LLM & Direct & 1.473 & 1.628 & 0.740 & 0.947 & 1.366 & 1.218 & 0.450 & 0.782 & 1.366 & 1.162 & 0.360 & 0.813 \\ 
\hline
(iv) ABL (w/o reflect, phy) & Iterative & 1.303 & 1.570 & 0.849 & 0.840 & 1.191 & 1.184 & 0.766 & 0.880 & 1.274 & 1.152 & 0.679 & 0.697 \\
(v) ABL (w/o design, phy) & Iterative & 1.306 & 1.381 & 0.667 & 0.677 & 1.198 & 1.186 & 0.611 & 0.704 & 1.230 & 1.164 & 0.422 & 0.677 \\ 
(vi) ABL (w/o phy) & Iterative & {1.278} & {1.316} & {0.903} & {0.973} & {1.182} & {1.181} & {0.796} & {0.944} & {1.193} & {1.151} & {0.729} & {0.893} \\
(vii) ABL (w/o llm) & Iterative & \multicolumn{2}{c}{1.659} & \multicolumn{2}{c|}{0.539} & \multicolumn{2}{c}{1.206} & \multicolumn{2}{c|}{0.644} & \multicolumn{2}{c}{1.206} & \multicolumn{2}{c}{0.557} \\
\hline
\textbf{Phythesis (Full)} & Iterative & \textbf{1.251} & \textbf{1.285} & \textbf{0.929} & \textbf{0.997} & \textbf{1.168} & \textbf{1.179} & \textbf{0.938} & \textbf{0.988} & \textbf{1.175} & \textbf{1.145} & \textbf{0.931} & \textbf{0.970} \\
\bottomrule
\end{tabular}
\end{table*}

\subsubsection{Design task}
We develop a set of design prompts for DC scene generation, containing 45 design requirements in three dimensions.
\textit{Regional profiles} cover tropical, dry, and temperate climate regions.
\textit{Computing capacities} include small-edge (50–100 servers), medium-cluster (1,000+ servers), and large-cloud (10,000+ servers).
\textit{Operational objective} here focuses only on minimizing PUE. 
These scenarios challenge Phythesis to balance semantic intent with physics-based industrial 3D scenes, testing its ability across various prompts.
We use a synthetic asset library that contains 10 different types of each asset introduced in Figure~\ref{fig:dtfile}.
We evaluate the performance of Phythesis through multiple backbone LLMs.
We employ two types of LLM for the generation task, i.e., instructive LLM (I-LM) and reasoning LLM (R-LM). Specifically, we use \texttt{GPT-4} as the I-LM and \texttt{o3-mini} as the R-LM.
All models serve as interchangeable backbones within the same framework.
We fix the LLM sampling parameters (temperature=0.7, $N=5$) and $K=5$ to ensure that the observed performance variances occur only from physics-guided logic rather than from model-specific capabilities. 

\subsubsection{Baselines}
We consider three baseline methods, including:
(i) \textbf{Random}, which initially explores DC designs through scene synthesis without leveraging physics priors or iterative refinements,
(ii) \textbf{EA} (i.e., evolutionary algorithm), which removes all LLM modules in Phythesis and adopts a population-based optimization method. It begins with random sampling and then mutates and selects candidate policies based on feedback from the physics engine, and
(iii) \textbf{Vanilla LLM}, which directly generates the semantic design candidates under the Design LLM without iterations.
We further include four ablation settings to isolate the contributions of our Phythesis components:
(iv) \textbf{ABL (w/o reflect, phy)}, which removes the Reflection LLM and asset optimization modules, adopting the same optimization operator as the EA baseline, 
(v) \textbf{ABL (w/o design, phy)}, which removes the Design LLM and the asset optimization modules, adopting random sampling for the initiation,
(vi) \textbf{ABL (w/o phy)}, which excludes asset optimization but retains both the Design LLM and Reflection LLM for design generation and refinement, and
(vii) \textbf{ABL (w/o llm)}, which keeps only asset optimization and removes the Design LLM and the Reflection LLM modules, adopting random sampling for the initiation.

\subsection{Experiment Results}
This section presents the experiment results of Phythesis on DC generative design. 
We assess synthetic scenes through two metrics.
\textit{Generation success rate (GSR)} measures the percentage of sampled design candidates that are valid in syntax and adhere to the constraints in Equation~(\ref{eq:problem}).
\textit{PUE} measures the sustainability objective. We calculate the best PUE result of all the designs generated throughout the iterations.
We compare Phythesis with baseline and ablation methods in terms of generation quality across scenarios and adaptability to different asset library sizes.

\subsubsection{Generation quality}
As shown in Table~\ref{tab:performance}, Phythesis achieves superior performance across all task dimensions compared with baselines and ablations, demonstrating the effectiveness of its physics-guided LLM synthesis framework.
Phythesis outperforms direct baselines by large margins. Compared with \textit{Random}, it achieves 240.3\%, 131.4\%, and 126.1\% higher GSR in all scenarios when using the R-LM backbone. Compared with \textit{Vanilla LLM}, Phythesis achieves up to 158.6\% relative increase in GSR (I-LM, large-cloud) and up to 21.1\% relative reduction in PUE (R-LM, small-edge).
In particular, Phythesis achieves PUE 0.283, 0.119, and 0.104 lower than \textit{Vanilla LLM} in the three scenarios, highlighting its efficiency in energy-aware synthesis. Regarding ablation settings,
disabling the Design LLM (cf. \textit{ABL (w/o design, phy)}) degrades GSR by 28.7\% in medium-cluster and 30.2\% in large-cloud with the R-LM backbone, confirming the necessity of the in-context physics priors.
Removing the Reflection LLM (cf. \textit{ABL (w/o reflect, phy)}) increases the PUE by 0.4\% in medium-cluster and 0.6\% in large-cloud when using R-LM, underscoring the role of physics feedback reflection.
Excluding physics-informed asset optimization (cf. \textit{ABL (w/o phy)}) leads to GSR drops by 2.4\%, 4.5\%, and 7.9\%, as well as PUE increase by 2.4\%, 0.2\%, and 0.5\%, across the three scenarios with the R-LM backbone, suggesting the value of physics-informed optimization for fine-tuning physical parameters.
Removing the LLM modules entirely (cf. \textit{ABL (w/o llm)})  results in 45.9\% GSR drop in small-edge and 42.6\% in large-cloud, and up to 0.374 PUE increase in small-edge, emphasizing the critical role of semantic reasoning from the Design and Reflection LLMs over pure optimization with random initialization.
In summary, the bi-level optimization flows mutually benefit each other to generate high-quality DC designs.

\begin{table}[t]
\centering
\caption{Performance comparison of different settings of Phythesis under different asset library scales. }
\label{tab:extensive_asset}
\vspace{-1em}
\begin{tabular}{l|cc|cc|cc}
\toprule
\multirow{2}{*}{\textbf{Method}} & 
\multicolumn{2}{c|}{\textbf{20 Assets}} & 
\multicolumn{2}{c|}{\textbf{30 Assets}} & 
\multicolumn{2}{c}{\textbf{50 Assets}} \\
\cmidrule(lr){2-7}
& PUE$\downarrow$ & GSR$\uparrow$ 
& PUE$\downarrow$ & GSR$\uparrow$
& PUE$\downarrow$ & GSR$\uparrow$ \\
\midrule
EA & 1.753 & 0.560 & 1.389 & 0.624 & 1.545 & 0.556 \\
w/o phy & 1.310 & 0.907 & 1.304 & 0.937 & 1.290 & 0.880 \\
w/o llm & 1.410 & 0.581 & 1.492 & 0.647 & 1.338 & 0.620 \\
\textbf{Phythesis} & \textbf{1.266} & \textbf{0.908} & \textbf{1.220} & \textbf{0.960} & \textbf{1.218} & \textbf{0.973} \\
\bottomrule
\end{tabular}
\end{table}

\subsubsection{Adaptability to extensive asset library}
To evaluate the scalability of Phythesis, we extend the original asset library to 20, 30, and 50 assets for each asset type, respectively. 
The enlarged asset libraries are designed to test whether the inference and search efficiency of different approaches can remain stable when the asset pool becomes more diverse and potentially more ambiguous. 
We compare \textit{EA}, \textit{ABL (w/o phy)}, and \textit{ABL (w/o llm)} with Phythesis on the above asset libraries. 
As shown in Table~\ref{tab:extensive_asset}, Phythesis still outperforms baselines when the asset library grows.
With 20, 30, and 50 assets, Phythesis reduces PUE relative to \textit{EA} by 27.8\%, 12.2\%, and 21.2\%, respectively, while maintaining the highest GSR across all settings. For the 50-asset library, Phythesis achieves a PUE of 1.218, which is 21.2\% lower than \textit{EA}, 5.6\% lower than \textit{ABL (w/o phy)}, and 9.0\% lower than  \textit{ABL (w/o llm)}. 
This confirms the advantage of physics-informed search when navigating a more diverse combinatorial space.

\subsection{Ablation Studies}
This section presents ablation studies to show the effectiveness of each design component in Phythesis. 
We isolate and compare individual modules against baseline and ablation settings to examine their contributions to initialization, optimization efficiency, and final performance. 

\begin{figure}[t]
    \centering
    \includegraphics[width=\linewidth]{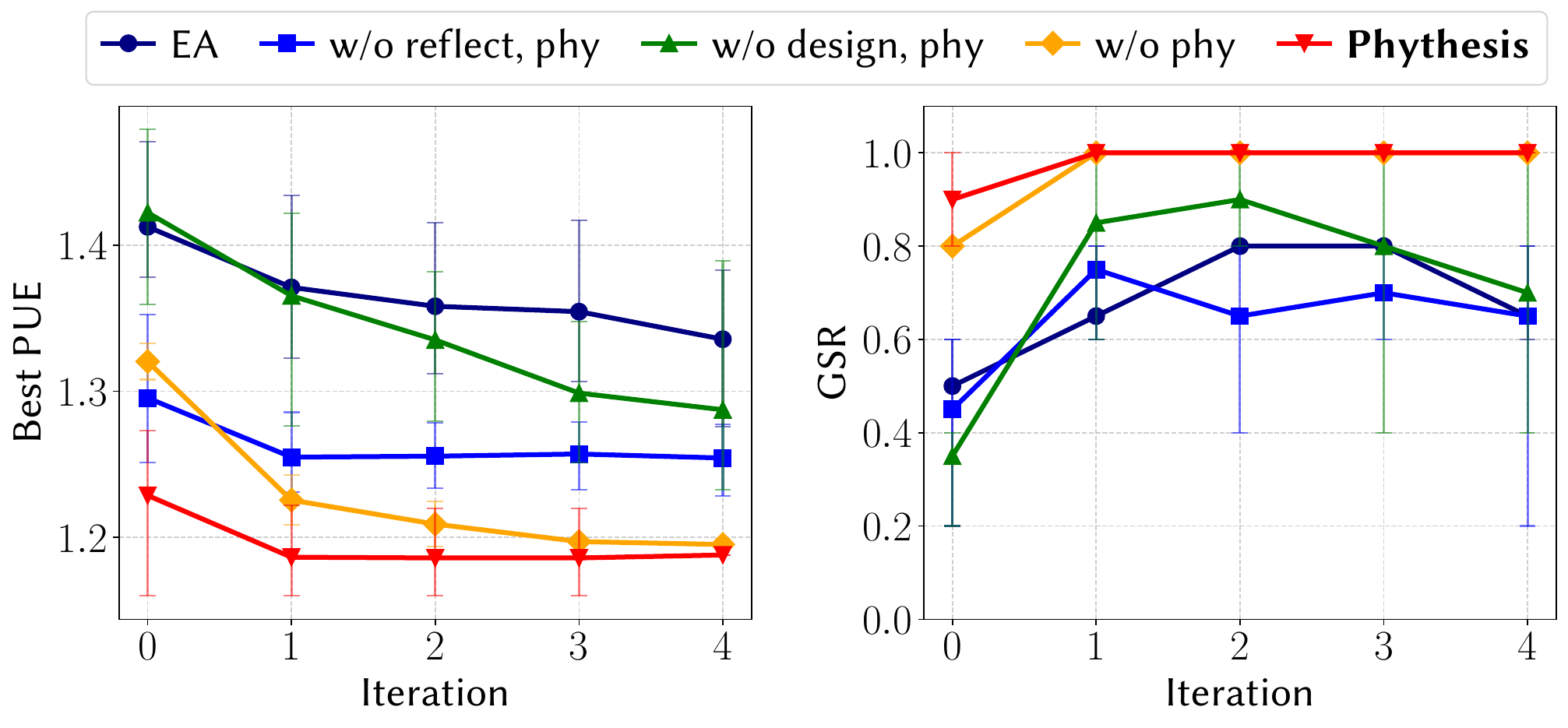}
    \vspace{-1em}
    \caption{Performance trends for DC designs, including \textit{EA}, ablation settings (iv)-(vi), and Phythesis. }
    \label{fig:trend}
\end{figure}

\vspace{-0.5em}
\subsubsection{Evolution trend and speed}

We analyze the evolutionary trajectories of a large-scale cloud scene synthesis using I-LM, by comparing Phythesis against \textit{EA} and ablations settings (iv)-(vi). Each experiment is repeated five times to ensure robustness.

Three key observations highlight the critical role of physics guidance in Phythesis’s design modules as shown in Figure~\ref{fig:trend}.
First, incorporating physics priors enables better initialization. \textit{ABL (w/o reflect, phy)} and Phythesis achieve an initial PUE more than 0.15 lower than \textit{EA} and \textit{ABL (w/o design, phy)}, attributed to the Design LLM. 
In contrast, \textit{EA} and \textit{ABL (w/o reflect, phy)} generate designs without prior knowledge, starting with more than 1.4 PUE at the first iteration. 
With the Design LLM for better initialization, \textit{ABL (w/o reflect, phy)} outperforms \textit{EA} by 7\% relative improvement in PUE.
This demonstrates the importance of physics priors in the Design LLM for feasible initialization.
In addition, physics-guided reflection accelerates optimization. 
Although \textit{ABL (w/o reflect, phy)} matches the initial performance of Phythesis, its PUE reduction stagnates after the first iteration, where cooling efficiency improves only 0.03 versus 0.11 of Phythesis with physics-guidance cycles. 
This LLM-driven optimization increases \textit{ABL (w/o design, phy)} in the reduction of PUE by nearly 100\% compared with \textit{EA} after 4 iterations, confirming that the physics feedback is vital for sustained evolution.
Finally, the physics-informed optimization module enables more precise and efficient convergence. 
\textit{ABL (w/o phy)} shows an initial PUE similar to \textit{ABL (w/o reflect, phy)}, but a much higher PUE than Phythesis, due to reliance on less directed search mechanisms. 
In comparison, Phythesis uses physics-informed optimization to refine designs more effectively, outperforming \textit{ABL (w/o phy)} by 15\% relative improvement in the final PUE and demonstrating faster overall optimization. 
Phythesis achieves a GSR of 0.85 in the first iteration, underscoring the importance of physics-informed optimization in ensuring stable generative design and optimization. In contrast, the GSR of other baselines shows noticeable fluctuations. \textit{ABL (w/o phy)} gains a slightly lower GSR than Phythesis, indicating that physics-informed optimization offers more stable and consistent improvements in generative design from the start.
These results underscore the significance of designed modules in faster evolutionary optimization and better industrial scene synthesis performance.

\begin{figure}[t]
    \centering
    \includegraphics[width=0.95\linewidth]{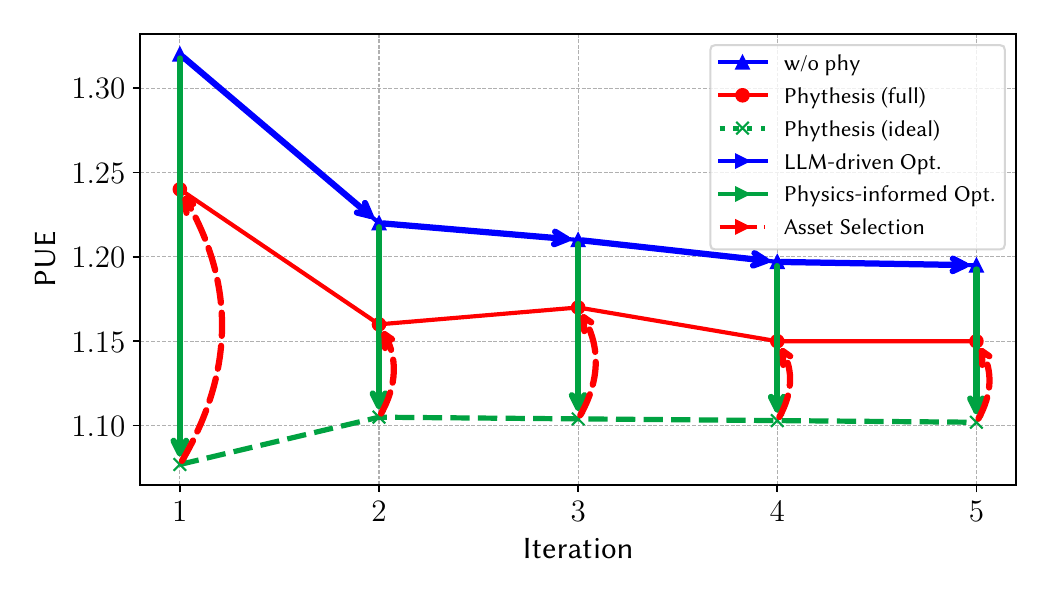}
    \vspace{-2em}
    \caption{The PUE trends and optimization flows of \textit{ABL (w/o phy)}, Phythesis (ideal), and Phythesis (full).}
    \label{fig:dual_loop}
\end{figure}

\vspace{-0.5em}
\subsubsection{Role of bi-level optimization}
We conduct an ablation study on the impact of our bi-level optimization strategy using the same large-cloud scene generation example above. 
Besides \textit{ABL (w/o phy)} and Phythesis, this experiment includes an extra variant where Phythesis generates the DC designs with the ideal asset parameters.
Figure~\ref{fig:dual_loop} illustrates the PUE trends across several variants over five iterations.
\textit{ABL (w/o phy)}, which omits physics-informed optimization, starts at a PUE of 1.320 and decreases gradually to around 1.195.
The complete implementation of our bi-level approach achieves the most consistent and efficient PUE reductions from 1.228 to 1.188.
This achievement is brought by the physics-informed optimization, which generates the scene with the ideal asset combination that starts at 1.077 PUE in the first iteration.
By selecting the nearest assets, Phythesis achieves lower PUE compared with ablation settings lacking physics-informed optimization. 
These results highlight the synergistic benefits of applying physics-informed guidance to the LLM-driven generative design and evolutionary optimization.

\begin{figure}[t]
    \centering
    \includegraphics[width=\linewidth]{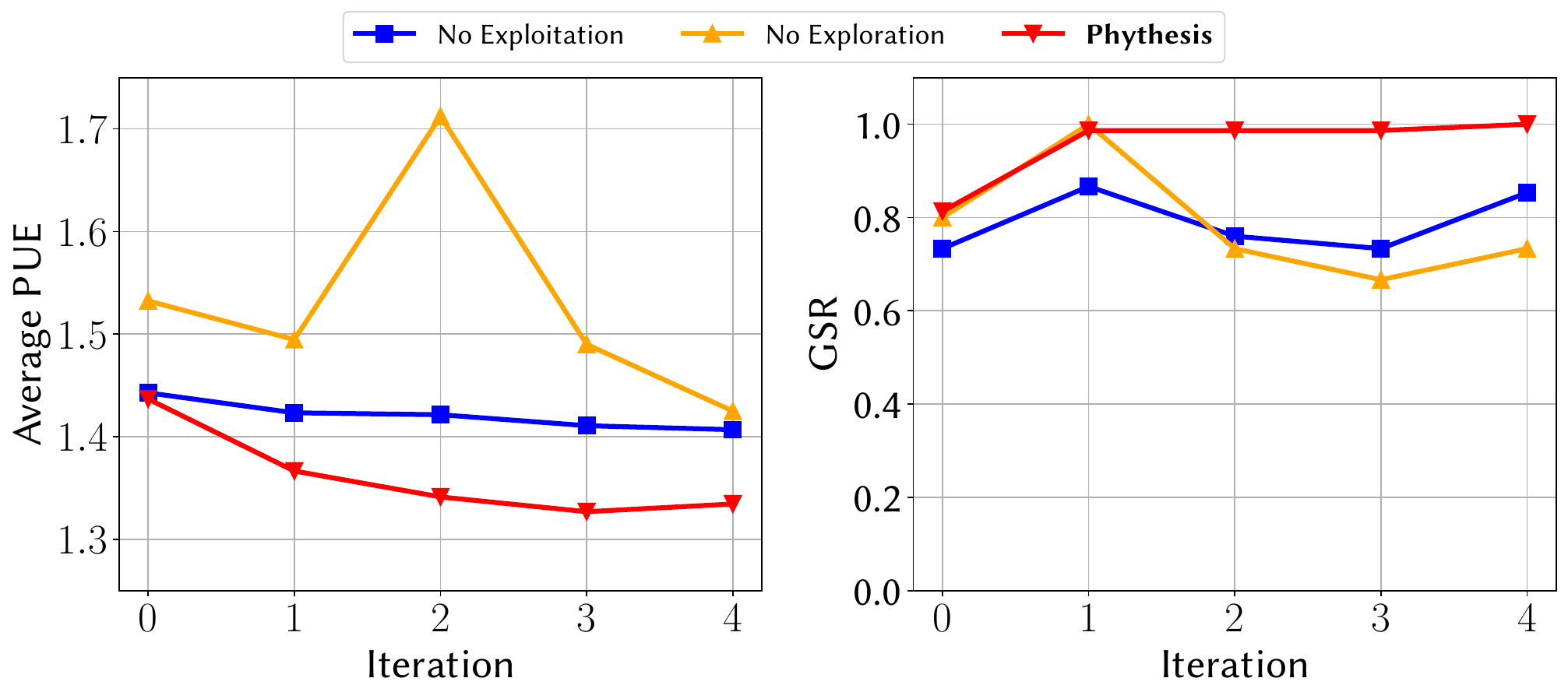}
    \vspace{-2em}
    \caption{GSR and PUE of generated DC designs under no-exploration, no-exploitation, and Phythesis across 5 iterations in the small-edge generation.}
    \vspace{-1.5em}
    \label{fig:sample_topk}
\end{figure}

\vspace{-0.5em}
\subsubsection{Exploration and exploitation}
To evaluate the impact of sampling size $N$ and top-$K$ selection in each iteration, we compare Phythesis with two baseline configurations using the small-edge scene generation task. First, we use $N=1,K=5$, representing a \textit{no-exploration} setting with minimal sampling diversity. Second, we use $N=5,K=0$, embodying a \textit{no-exploitation} setting without top designs in the LLM-driven optimization.
Figure~\ref{fig:sample_topk} presents the GSR and average PUE for DC designs generated across these configurations, together with the full Phythesis. 
The figure on the left illustrates the average PUE trends, where Phythesis achieves the lowest and most stable values at 1.327 average PUE, underscoring its efficiency in optimizing energy consumption. The no-exploration Phythesis decreases moderately from 1.431 to 1.411 average PUE, whereas the no-exploitation version starts with a lower initial PUE.
As shown in the figure on the right, Phythesis without exploration experiences fluctuations in GSR over iterations. In contrast, the no-exploitation baseline starts at 0.8 but maintains the same GSR during the iterations. Phythesis starts at 0.813 GSR and maintains a consistently high GSR through iterations, demonstrating the effectiveness of balanced exploration and exploitation.
These results suggest that the exploration by LLM sampling and the exploitation of top designs by in-context learning are critical for the LLM-driven optimization in Phythesis.

\vspace{-0.5em}
\section{Case Study on Sustainability Impact}
\label{sec:case}

\begin{figure}[t]
    \centering
    \includegraphics[width=\linewidth]{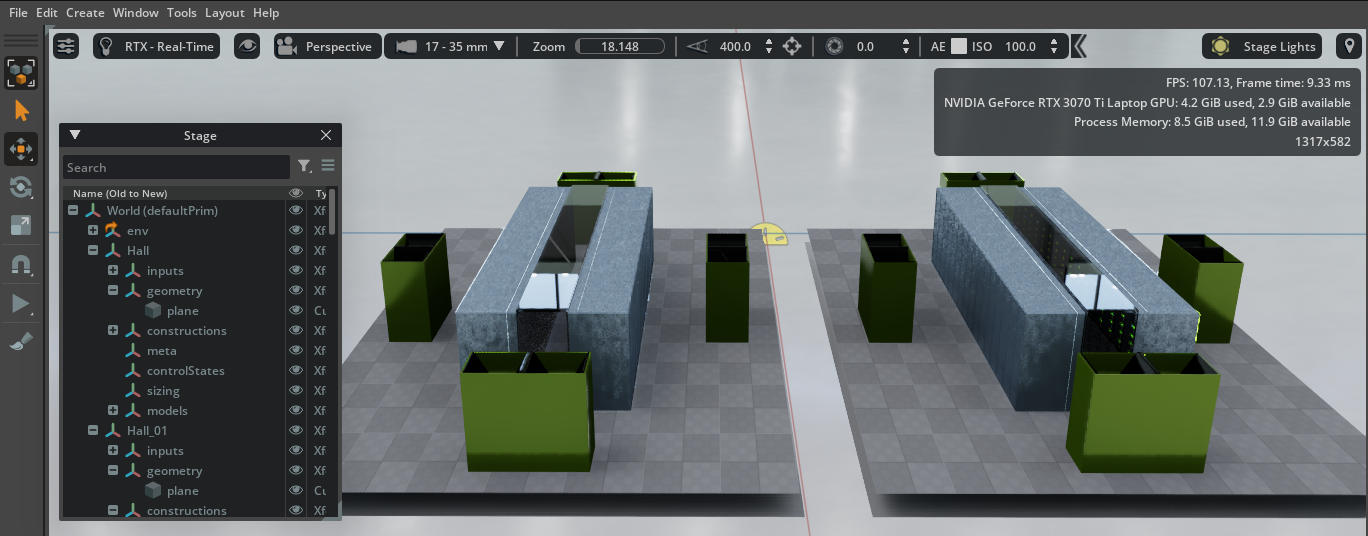}
    \\
    \vspace{0.5em}
    \includegraphics[width=\linewidth]{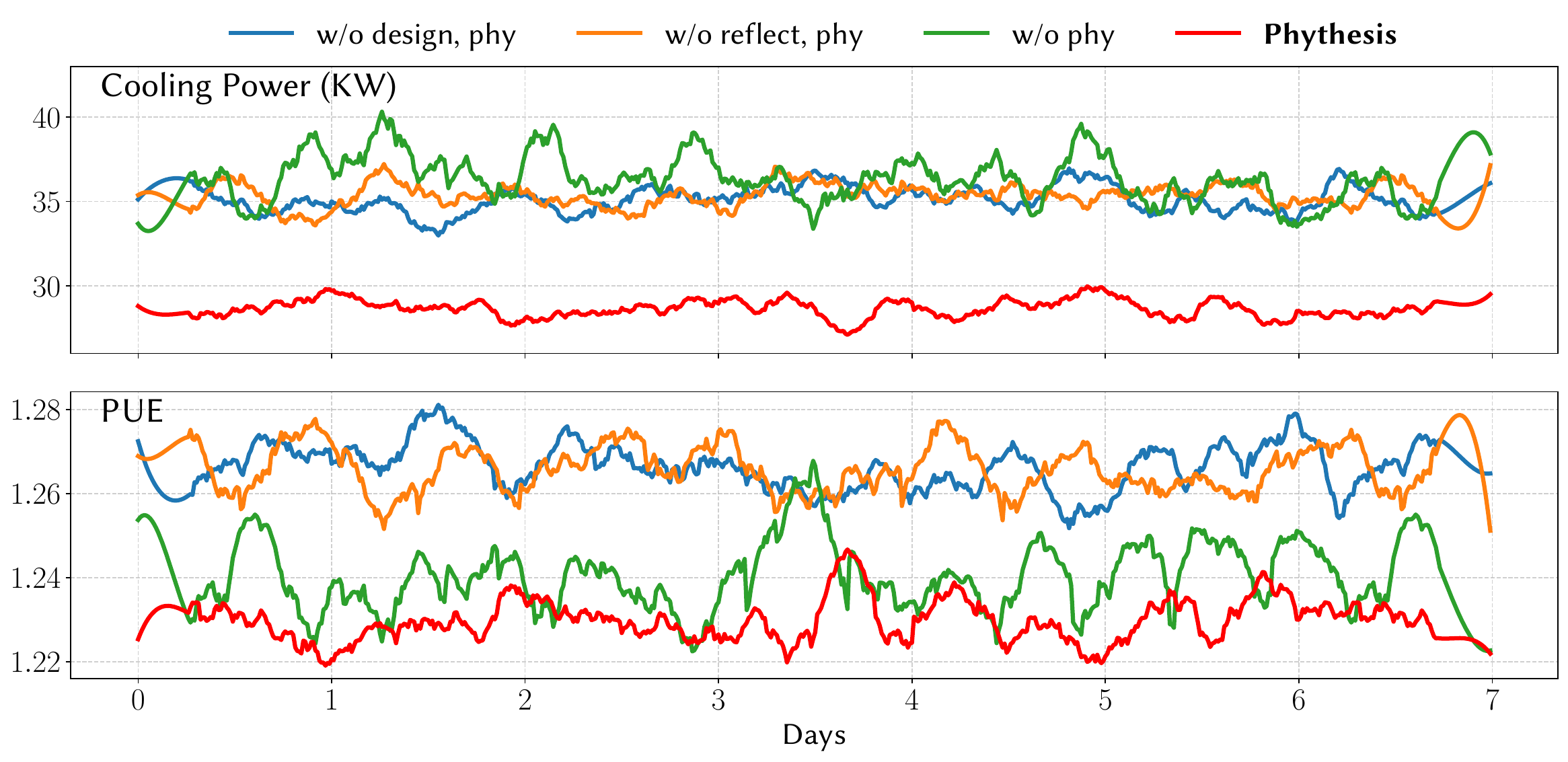}
    \vspace{-2.5em}
    \caption{Visualization of the generated 3D server rooms by Phythesis (Top), and simulated cooling power and PUE trajectories comparing ablation methods with Phythesis (Bottom).}
    \vspace{-1.5em}
    \label{fig:case_study_composite}
\end{figure}

To demonstrate the sustainability impact of Phythesis, this section presents a case study on a small-edge DC generated by the following prompt: \textit{"Design a data center suitable for small-edge deployments in a tropical region, with a computational capacity between 4 and 8 PFLOPS. The design must prioritize energy efficiency to achieve a PUE below 1.3."}
Figure~\ref{fig:case_study_composite} shows the generated 3D server rooms.

We conduct a week-long simulation under the historical weather of Singapore from~\cite{crawley2001energyplus} and random workload ratios with the supply air temperature ranging from 12$^\circ$C to 22$^\circ$C. Figure~\ref{fig:case_study_composite} illustrates the cooling power and PUE trajectories under different baselines and Phythesis. 
Phythesis achieves a 6.2\% PUE reduction compared to the industry standard of 1.3. 
In particular, Phythesis maintains a more stable PUE curve and achieves a consistent 0.01–0.02 reduction in PUE relative to the best-performing \textit{ABL (w/o phy)}, corresponding to an improvement from 1.25 to 1.23 approximately on average. 
Although such a reduction may appear marginal, it translates into around 0.6 GWh of weekly energy savings, or 31.2 GWh annually for a large-scale DC deployment. The impact is amplified when compared with \textit{ABL (w/o design, phy)} and \textit{ABL (w/o reflect, phy)}, where Phythesis’s cooling systems save 0.7 GWh weekly, translating to 38.5 GWh annually. This savings is equivalent to powering all households in a city of 5 million people, such as Singapore, for nearly two days~\cite{ema2025ses}. 
This leap in efficiency marks a significant sustainability achievement potential by DC design.

\vspace{-0.5em}
\section{Discussions}
\label{sec:discuss}
Our implementation fundamentally depends on the simulator's predefined boundaries (i.e., simulation fidelity and simulatable components).
Modeling intricate layouts of cabling, piping, and power distribution requires extensive extra work, which is beyond the scope of this paper. 
To elevate Phythesis from an early-stage design framework into an industrial-grade tool, we now discuss its limitations and future directions.
First, in this work, we only consider PUE as the only objective for energy-efficient DC design.
This energy-aware focus may suit the design of a prefabricated modular DC~\cite{HuaweiPrefabricatedModular}, which simplifies these elements and prioritizes energy efficiency.
Future work could consider multi-objective optimization, such as minimizing water usage \cite{shumba2025water}, carbon emissions \cite{qi2023mosaic}, or construction costs \cite{van2020experiences}. 
Besides, the simulator in this paper employs parameterized, component-level abstractions instead of full thermodynamic simulation.
Future work could leverage multimodal encoders to understand flow fields, enabling semantic comprehension of high-dimensional simulation results~\cite{cao2025transforming}.
Second, our experiments involve typical ITE and cooling facilities found in today's air-cooled DCs.
Future work can broaden SimReady assets to cover diverse cooling technologies (e.g., direct-to-chip liquid, rear-door heat exchangers) tailored to heterogeneous ITE, and to model regional resource conditions for siting-aware DC design~\cite{cao2025transforming}.
Third, our evolutionary optimization process currently does not provide a formal convergence guarantee. 
Besides, the "optimize-then-select" strategy is effective but may converge to local optima due to non-linear asset interactions.
Extending Phythesis to address intricate objectives may amplify uncertainties in LLM outputs, thus challenging the convergence. Rigorous analysis of optimality, convergence, hallucination risks, alongside the development of corresponding finetuning strategies~\cite{huang2025survey}, constitutes an important future direction.


\vspace{-0.5em}
\section{Conclusion}
\label{sec:conclude}
This paper presented Phythesis, a physics-guided evolutionary scene synthesis framework for generative DC design.
The evolutionary process employs a bi-level optimization strategy.
The LLM-driven optimization incorporates physics prior knowledge and simulation feedback to mitigate the LLM's hallucinations and provide better initialization and fast iteration.
The physics-informed optimization further optimizes asset selection through the use of a differentiable physics-based engine.
We ran experiments on three different scales of the DC design tasks. 
Our approach outperforms other baseline methods and ablation settings in terms of generation success rate and the expected energy efficiency of the resulting design.
We further validated the sustainability impact of our approach through a DC design case.
Phythesis provides a
blueprint to harmonize the rapid innovation of energy-efficient infrastructures.

\balance
\bibliographystyle{ACM-Reference-Format}
\bibliography{sample-base}

@String{Computing = "Computing" }

@String{Computer = "{IEEE} Computer" }

@String{Springer = "Springer-Verlag" }

@article{feng2023layoutgpt,
  title={Layoutgpt: Compositional visual planning and generation with large language models},
  author={Feng, Weixi and Zhu, Wanrong and Fu, Tsu-jui and Jampani, Varun and Akula, Arjun and He, Xuehai and Basu, Sugato and Wang, Xin Eric and Wang, William Yang},
  journal={Advances in Neural Information Processing Systems},
  volume={36},
  pages={18225--18250},
  year={2023}
}

@article{sun20233d,
  title={3d-gpt: Procedural 3d modeling with large language models},
  author={Sun, Chunyi and Han, Junlin and Deng, Weijian and Wang, Xinlong and Qin, Zishan and Gould, Stephen},
  journal={arXiv preprint arXiv:2310.12945},
  year={2023}
}

@article{zhang2024scenepalette,
  title={ScenePalette: Contextually Exploring Object Collections Through Multiplex Relations in 3D Scenes},
  author={Zhang, Shao-Kui and Xie, Wei-Yu and Wang, Chen and Zhang, Song-Hai},
  journal={Journal of Computer Science and Technology},
  volume={39},
  number={5},
  pages={1180--1192},
  year={2024},
  publisher={Springer}
}

@inproceedings{qi2018human,
  title={Human-centric indoor scene synthesis using stochastic grammar},
  author={Qi, Siyuan and Zhu, Yixin and Huang, Siyuan and Jiang, Chenfanfu and Zhu, Song-Chun},
  booktitle={Proceedings of the IEEE Conference on Computer Vision and Pattern Recognition},
  pages={5899--5908},
  year={2018}
}

@inproceedings{gao2024graphdreamer,
  title={Graphdreamer: Compositional 3d scene synthesis from scene graphs},
  author={Gao, Gege and Liu, Weiyang and Chen, Anpei and Geiger, Andreas and Sch{\"o}lkopf, Bernhard},
  booktitle={Proceedings of the IEEE/CVF Conference on Computer Vision and Pattern Recognition},
  pages={21295--21304},
  year={2024}
}

@inproceedings{xu2024penetrative,
  title={Penetrative ai: Making llms comprehend the physical world},
  author={Xu, Huatao and Han, Liying and Yang, Qirui and Li, Mo and Srivastava, Mani},
  booktitle={Proceedings of the 25th International Workshop on Mobile Computing Systems and Applications},
  pages={1--7},
  year={2024}
}

@article{huang2025survey,
  title={A survey on hallucination in large language models: Principles, taxonomy, challenges, and open questions},
  author={Huang, Lei and Yu, Weijiang and Ma, Weitao and Zhong, Weihong and Feng, Zhangyin and Wang, Haotian and Chen, Qianglong and Peng, Weihua and Feng, Xiaocheng and Qin, Bing and others},
  journal={ACM Transactions on Information Systems},
  volume={43},
  number={2},
  pages={1--55},
  year={2025},
  publisher={ACM New York, NY}
}

@article{xia2024scenegenagent,
  title={Scenegenagent: Precise industrial scene generation with coding agent},
  author={Xia, Xiao and Zhang, Dan and Liao, Zibo and Hou, Zhenyu and Sun, Tianrui and Li, Jing and Fu, Ling and Dong, Yuxiao},
  journal={arXiv preprint arXiv:2410.21909},
  year={2024}
}

@inproceedings{de2024llmr,
  title={Llmr: Real-time prompting of interactive worlds using large language models},
  author={De La Torre, Fernanda and Fang, Cathy Mengying and Huang, Han and Banburski-Fahey, Andrzej and Amores Fernandez, Judith and Lanier, Jaron},
  booktitle={Proceedings of the 2024 CHI Conference on Human Factors in Computing Systems},
  pages={1--22},
  year={2024}
}

@inproceedings{li2023chattwin,
  title={ChatTwin: Toward Automated Digital Twin Generation for Data Center via Large Language Models},
  author={Li, Minghao and Wang, Ruihang and Zhou, Xin and Zhu, Zhaomeng and Wen, Yonggang and Tan, Rui},
  booktitle={Proceedings of the ACM Conference on Systems for Energy-Efficient Built Environments (BuildSys)},
  pages={208--211},
  year={2023}
}

@article{jiang2024eplus,
  title={EPlus-LLM: A Large Language Model-Based Computing Platform for Automated Building Energy Modeling},
  author={Jiang, Gang and Ma, Zhihao and Zhang, Liang and Chen, Jianli},
  journal={Applied Energy},
  volume={367},
  pages={123431},
  year={2024},
  publisher={Elsevier}
}

@inproceedings{ma2024llm,
  title={LLM and simulation as bilevel optimizers: a new paradigm to advance physical scientific discovery},
  author={Ma, Pingchuan and Wang, Tsun-Hsuan and Guo, Minghao and Sun, Zhiqing and Tenenbaum, Joshua B and Rus, Daniela and Gan, Chuang and Matusik, Wojciech},
  booktitle={Proceedings of the 41st International Conference on Machine Learning},
  pages={33940--33962},
  year={2024}
}

@article{holt2024automatically,
  title={Automatically Learning Hybrid Digital Twins of Dynamical Systems},
  author={Holt, Samuel and Liu, Tennison and van der Schaar, Mihaela},
  journal={arXiv preprint arXiv:2410.23691},
  year={2024}
}

@misc{nvidia_simready_assets,
  author = {{NVIDIA Corporation}},
  title = {SimReady Assets},
  year = {2023},
  url = {https://developer.nvidia.com/omniverse/simready-assets},
  howpublished = {NVIDIA Omniverse Developer Portal}
}

@article{liu2023reason,
  title={Reason for future, act for now: A principled framework for autonomous llm agents with provable sample efficiency},
  author={Liu, Zhihan and Hu, Hao and Zhang, Shenao and Guo, Hongyi and Ke, Shuqi and Liu, Boyi and Wang, Zhaoran},
  journal={arXiv preprint arXiv:2309.17382},
  year={2023}
}

@article{McKinsey2024,
  author       = {Bhargs Srivathsan and Marc Sorel and Pankaj Sachdeva},
  title        = {AI power: Expanding data center capacity to meet growing demand},
  journal      = {McKinsey \& Company},
  year         = {2024},
  month        = {October},
  url          = {https://www.mckinsey.com/industries/technology-media-and-telecommunications/our-insights/ai-power-expanding-data-center-capacity-to-meet-growing-demand}
}

@article{forbestechcouncil2024,
  author = {{Forbes Technology Council}},
  title = {Data Centers: 18 Challenges And Solutions On The Horizon},
  journal = {Forbes},
  year = {2024},
  url = {https://www.forbes.com/councils/forbestechcouncil/2024/12/19/data-centers-18-challenges-and-solutions-on-the-horizon/},
}

@article{shehabi2011data,
  title={Data center design and location: Consequences for electricity use and greenhouse-gas emissions},
  author={Shehabi, Arman and Masanet, Eric and Price, Hillary and Horvath, Arpad and Nazaroff, William W},
  journal={Building and Environment},
  volume={46},
  number={5},
  pages={990--998},
  year={2011},
  publisher={Elsevier}
}

@article{fakhim2011cooling,
  title={Cooling solutions in an operational data centre: A case study},
  author={Fakhim, Babak and Behnia, M and Armfield, SW and Srinarayana, N},
  journal={Applied thermal engineering},
  volume={31},
  number={14-15},
  pages={2279--2291},
  year={2011},
  publisher={Elsevier}
}

@article{cai2024towards,
  title={Towards energy-efficient data centers: A comprehensive review of passive and active cooling strategies},
  author={Cai, Senhong and Gou, Zhonghua},
  journal={Energy and Built Environment},
  year={2024},
  publisher={Elsevier}
}

@article{crawley2001energyplus,
  title={EnergyPlus: creating a new-generation building energy simulation program},
  author={Crawley, Drury B and Lawrie, Linda K and Winkelmann, Frederick C and Buhl, Walter F and Huang, Y Joe and Pedersen, Curtis O and Strand, Richard K and Liesen, Richard J and Fisher, Daniel E and Witte, Michael J and others},
  journal={Energy and buildings},
  volume={33},
  number={4},
  pages={319--331},
  year={2001},
  publisher={Elsevier}
}

@online{actiflow2025cfd,
  author = {Actiflow},
  title = {{CFD: The Truth and the Tales}},
  year = {2018},
  url = {https://actiflow.com/cfd-the-truth-and-the-tales-2/},
  urldate = {2018-08-11}
}

@misc{GB50174-2017,
  title     = {Data Center Design Specification},
  number    = {GB 50174-2017},
  year      = {2017},
  author = {Standardization Administration of China},
  url       = {https://baigongbao.oss-cn-beijing.aliyuncs.com/2020/10/14/pd3b2HC8QR.pdf},
  language  = {English (translated from Chinese)}
}

@misc{uptime_tier_topology,
  author       = {{Uptime Institute}},
  title        = {Tier Standard: Topology},
  year         = {2018},
  howpublished = {\url{https://uptimeinstitute.com/resources/asset/tier-standard-topology}},
}

@inproceedings{shumba2025water,
  title={A water efficiency dataset for African data centers},
  author={Shumba, Noah and Tshekiso, Opelo and Li, Pengfei and Fanti, Giulia and Ren, Shaolei},
  booktitle={Proceedings of the ACM SIGCAS/SIGCHI Conference on Computing and Sustainable Societies},
  pages={453--460},
  year={2025}
}

@inproceedings{qi2023mosaic,
  title={MOSAIC: A Multi-Objective Optimization Framework for Sustainable Datacenter Management},
  author={Qi, Sirui and Milojicic, Dejan and Bash, Cullen and Pasricha, Sudeep},
  booktitle={2023 IEEE 30th International Conference on High Performance Computing, Data, and Analytics (HiPC)},
  pages={51--60},
  year={2023},
  organization={IEEE}
}

@article{cao2025transforming,
  title={Transforming Future Data Center Operations and Management via Physical AI},
  author={Cao, Zhiwei and Li, Minghao and Lin, Feng and Jia, Jimin and Wen, Yonggang and Yin, Jianxiong and See, Simon},
  journal={arXiv preprint arXiv:2504.04982},
  year={2025}
}

@inproceedings{van2020experiences,
  title={Experiences and learned lessons from an air free-cooled tropical data center testbed},
  author={Van Le, Duc and Liu, Yingbo and Wang, Rongrong and Tan, Rui and Ngoh, Lek Heng},
  booktitle={Proceedings of the 7th ACM International Conference on Systems for Energy-Efficient Buildings, Cities, and Transportation},
  pages={160--169},
  year={2020}
}

@article{du2024text2bim,
  title={Text2BIM: Generating building models using a large language model-based multi-agent framework},
  author={Du, Changyu and Esser, Sebastian and Nousias, Stavros and Borrmann, Andr{\'e}},
  journal={arXiv preprint arXiv:2408.08054},
  year={2024}
}

@misc{ema2025ses,
  author = {Energy Market Authority},
  title = {Singapore Energy Statistics Chapter 3: Energy Consumption},
  year = {2025},
  url = {https://www.ema.gov.sg/resources/singapore-energy-statistics/chapter3},
}

@article{stull2011wet,
  title={Wet-bulb temperature from relative humidity and air temperature},
  author={Stull, Roland},
  journal={Journal of applied meteorology and climatology},
  volume={50},
  number={11},
  pages={2267--2269},
  year={2011}
}

@techreport{ocp2023ready,
  title={{OCP-Ready Data Centers: Program Mission, Methodology and Case Studies}},
  author={{Open Compute Project}},
  institution={Open Compute Project Foundation},
  year={2023},
  type={Technical Report},
  note={Rev. 1.0},
  url={https://www.opencompute.org/documents/ocp-ready-data-centers-program-mission-methodology-and-case-studies-rev-1-0-pdf}
}

@techreport{HuaweiPrefabricatedModular,
    author = {{Huawei Technologies Co., Ltd.}},
    title = {Prefabricated Modular Data Center White Paper},
    institution = {Huawei},
    year = {2023},
    url = {https://digitalpower.huawei.com/upload-pro/index/index/Prefabricated-Modular-Data-Center-White-Paper.pdf},
    note = {Accessed: 2023}
}

@article{forrest1996genetic,
  title={Genetic algorithms},
  author={Forrest, Stephanie},
  journal={ACM computing surveys (CSUR)},
  volume={28},
  number={1},
  pages={77--80},
  year={1996},
  publisher={ACM New York, NY, USA}
}

\clearpage

\appendix

\nobalance
\section{An Exploratory Comparison with Open-Source Industry Standards}
\label{sec:ocp_comparison}
To validate Phythesis's performance against industry benchmarks, we conduct a comparative analysis with Open Compute Project (OCP)-Ready\texttrademark{} DCs, which represent state-of-the-art designs optimized for energy efficiency~\cite{ocp2023ready}. 
The OCP documentation provides high-level settings such as critical IT power, design PUE, and location, but detailed facility parameters, including specific cooling configurations, rack layouts, and operational setpoints, are proprietary.
Furthermore, without calibration on real-world operational data, our simulator cannot fully replicate actual facility conditions. These constraints prevent a rigorous, controlled comparison.
Thus, we conduct the following exploratory comparison to validate Phythesis's ability to generate designs that are competitive with open-source industry standards under similar environmental and operational conditions. 

\begin{table}[h]
\centering
\caption{Design specifications of open-source OCP-Ready\texttrademark{} DCs. Annual average outdoor air wet-bulb temperature and humidity ratio are calculated from historical weather data.}
\vspace{-1em}
\label{tab:ocp-ocp2}
\begin{tabular}{@{} l | c c @{}}
\toprule
\textbf{OCP-Ready\texttrademark{} DC}  & \textbf{JAK2} & \textbf{Stockton 1}\\
\midrule
Solution Provider            & SpaceDC & Nautilus DT \\
Country                      & Indonesia & USA \\
Cooling Method               & Air-Cooled & Air-Cooled \\
IT Power (MW)                & 1.4 & 7.0 \\
Design PUE                   & 1.50 & 1.20 \\
Annualized PUE               & 1.50 & N/A \\
\midrule
Location                     & Jakarta & Stockton \\
Average Wet-bulb Temp. (\textdegree{}C) & 31.7 & 11.6 \\
Average Humidity Ratio           & 0.018 & 0.007 \\
\bottomrule
\end{tabular}
\end{table}

\subsection{Environmental Condition Modeling}
We specifically focus on air-cooled facilities to provide a consistent basis for comparison, as these systems face similar thermal design challenges and operate within comparable environmental constraints.
We select two representative OCP facilities with air-cooled systems operating in distinct climate zones: (1) SpaceDC's JAK2 facility in tropical-climate Jakarta, Indonesia, and (2) Nautilus DT's Stockton 1 facility in California's hot-summer Mediterranean climate~\cite{ocp2023ready}.
These selections enable us to evaluate Phythesis's design optimization capabilities across diverse environmental conditions while maintaining methodological consistency. The open-source specifications and calculated conditions of these two facilities are presented in Table~\ref{tab:ocp-ocp2}. 

\begin{figure}[t]
  \centering
  \begin{subfigure}[b]{\linewidth}
    \centering
    \includegraphics[width=\linewidth]{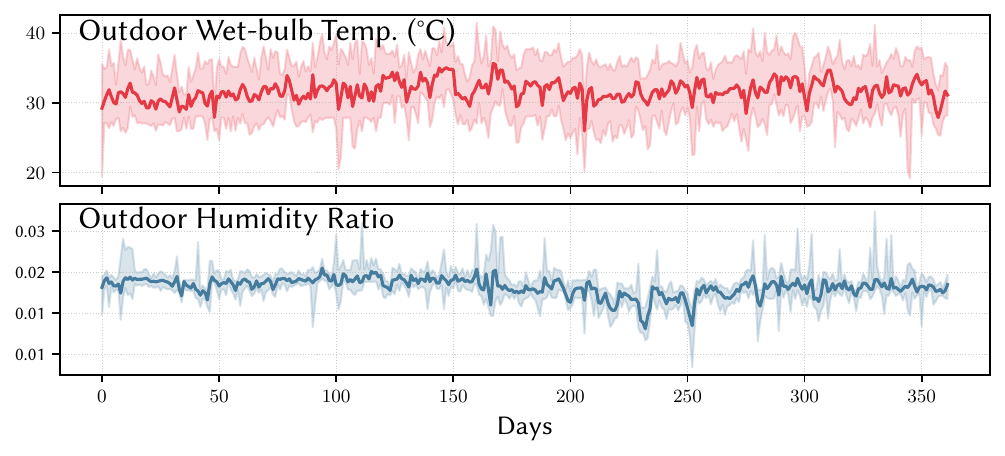}
    \vspace{-2em}
    \caption{JAK2}
    \label{fig:ocp_od_indo}
  \end{subfigure}
    \vspace{2em}
  \begin{subfigure}[b]{\linewidth}
    \centering
    \includegraphics[width=\linewidth]{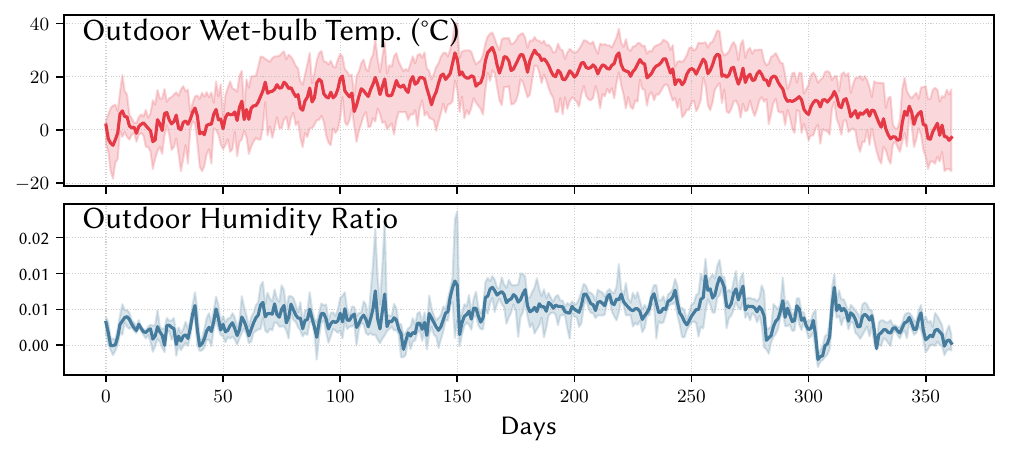}
    \vspace{-2em}
    \caption{Stockton 1}
    \label{fig:ocp_od_ca}
  \end{subfigure}
  \vspace{-4em}
  \caption{Annual outdoor wet-bulb temperature and air humidity ratio of JAK2 (Top) and Stockton 1 (Bottom).}
  \label{fig:ocp_od}
\end{figure}

We evaluate Phythesis-generated designs under identical external environmental conditions derived from historical weather data of the OCP facilities. External factors are calculated using EnergyPlus Weather (EPW) files\footnotemark{} for each respective location. The EPW files provide historical weather data, including dry-bulb temperature, dew-point temperature, relative humidity, and atmospheric pressure in time series. We employ the Stull approximation~\cite{stull2011wet} to calculate the outdoor air wet-bulb temperature and humidity ratio from the dry-bulb temperature and relative humidity extracted from the EPW files.
The wet-bulb temperature is calculated using the Stull approximation as:
\begin{align}
T_{wb} = &\, T \cdot \arctan(0.152\sqrt{RH + 8.314}) + \arctan(T + RH) \nonumber \\
&- \arctan(RH - 1.676) + 0.00392 \cdot RH^{1.5} \cdot \arctan(0.023RH) \nonumber \\
&- 4.686,
\end{align}
where $T_{wb}$ is the wet-bulb temperature in \textdegree{}C, $T$ is the dry-bulb temperature in \textdegree{}C, and $RH$ is the relative humidity in percentage.
The outdoor air humidity ratio is computed using standard psychrometric relations as:
\begin{equation}
w = 0.622 \cdot \frac{P_v}{P - P_v},
\end{equation}
where $P_v$ is the partial vapor pressure calculated as $RH \cdot P_{ws}/100$, $P_{ws}$ is the saturation vapor pressure at the given temperature, and $P$ is the atmospheric pressure.
Figure~\ref{fig:ocp_od} presents the annual profiles of the outdoor wet-bulb temperature and humidity ratio for each DC location (i.e., Jakarta and Stockton).
\footnotetext{EPW files were obtained from the Ladybug Tools EPW database at \url{https://www.ladybug.tools/epwmap}.}

\begin{figure*}[t]
  \centering
  \begin{subfigure}[b]{0.48\linewidth}
    \centering
    \includegraphics[width=\linewidth]{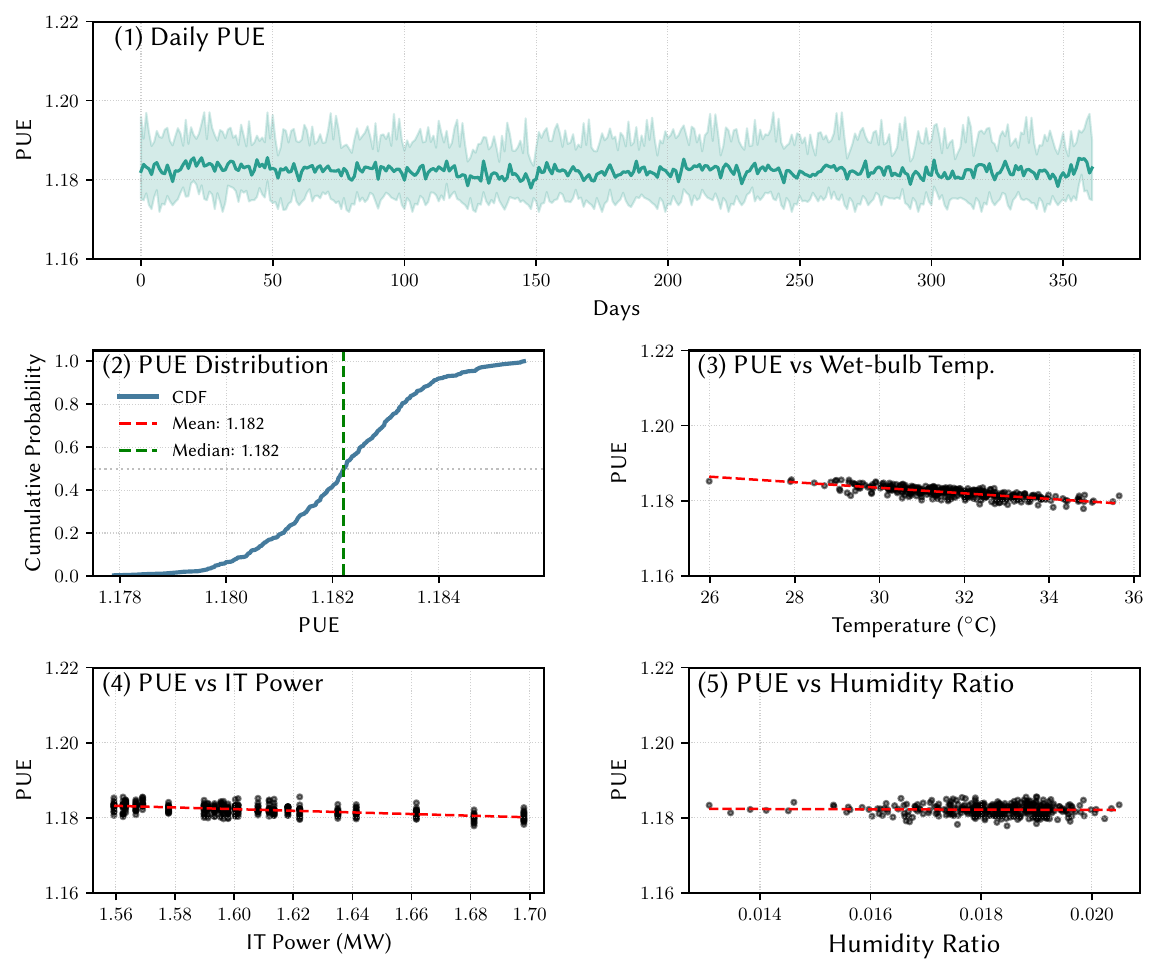}
    \vspace{-2em}
    \caption{JAK2}
    \label{fig:ocp_pue_indo}
  \end{subfigure}
  \hfill
  \begin{subfigure}[b]{0.48\linewidth}
    \centering
    \includegraphics[width=\linewidth]{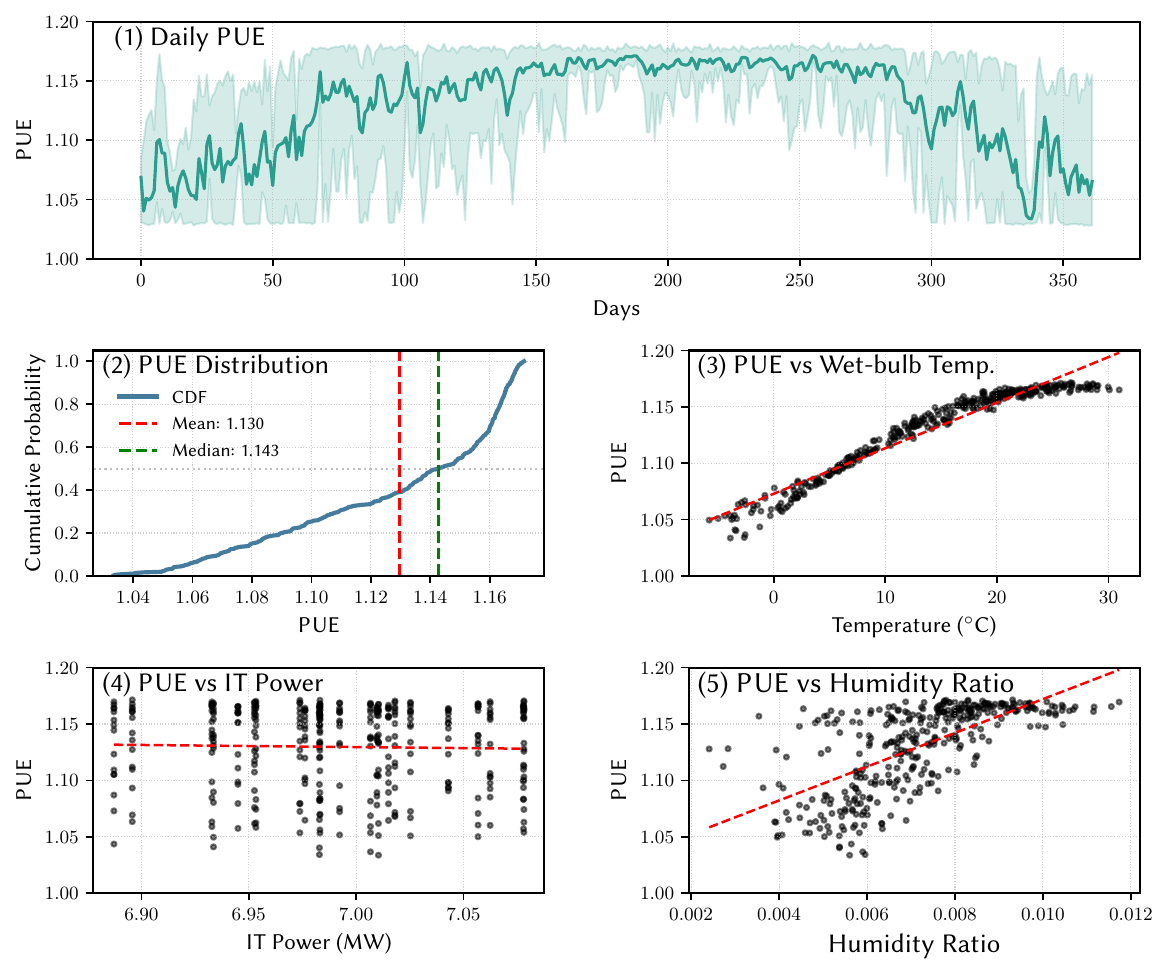}
    \vspace{-2em}
    \caption{Stockton 1}
    \label{fig:ocp_pue_ca}
  \end{subfigure}
  \caption{Annual simulation result of JAK2 (Left) and Stockton 1 (Right). We analyze the PUE trend and distributions and their relevant factors (i.e., outdoor temperature, humidity ratio, and IT power) of each generated design by Phythesis. }
  \label{fig:ocp_pue}
\end{figure*}

\subsection{Design Generation \& Performance Analysis}
To demonstrate Phythesis's capability to interpret real-world requirements, we used natural language prompts describing the core specifications of the JAK2 and Stockton 1 facilities to generate the corresponding designs. These prompts, shown in Table~\ref{tab:ocp_prompts}, encapsulate high-level goals such as location, IT capacity, rack layout, and thermal constraints. Phythesis then automatically translated these requirements into DC SimReady scenes.
The simulation of each generated design includes a full annual cycle with a 1-hour resolution, capturing seasonal variations in cooling demands.

\begin{table}[t]
\centering
\caption{Natural language prompts for generative design of Phythesis. These prompts describe the real-world design specifications of JAK2 and Stockton 1. }
\label{tab:ocp_prompts}
\begin{tabular}{@{}l>{\raggedright\arraybackslash}p{0.8\linewidth}@{}}
\toprule
\textbf{DC} & \textbf{Design Prompt} \\
\midrule
\textbf{JAK2} & Help me design an energy efficient data center in Jakarta. The total IT power should be at 1.4 MW (600-800 racks). Should at least 2 halls. Each with 300-400 racks. Please add enough cooling facilities to keep each zone temperature under 30. \\
\addlinespace
\textbf{Stockton 1} & Help me design an energy efficient data center in Stockton, California. The total IT power should be at 7 MW (3000-4000 racks). Should at least 10 halls. Each with 300-400 racks. Please add enough cooling facilities to keep each zone temperature under 30. \\
\bottomrule
\end{tabular}
\end{table}

We first analyze the JAK2 facility in Jakarta's tropical climate. As shown in Figure~\ref{fig:ocp_od}(a), the outdoor conditions exhibit consistently high temperatures and humidity throughout the year, with an annual average wet-bulb temperature of $\SI{31.71}{\celsius}$ and humidity ratio of 0.0182. These conditions present significant challenges for air-cooled operations.
The Phythesis-generated design achieves an average PUE of 1.182, representing a 21.2\% improvement over the standard of 1.5, as reported in Table~\ref{tab:ocp-ocp2}. As shown in Figure~\ref{fig:ocp_pue}(a), the design maintains an average IT power of 1.61 MW with cooling power consumption of 0.29 MW. The relatively stable tropical climate contributes to consistent PUE performance throughout the year. Notably, PUE decreases as IT power increases, a trend consistent with thermodynamic principles where higher IT loads improve the ratio of useful work to total energy consumption. This behavior validates the physical coherence of the generated design.

The Stockton facility operates under totally different conditions. As shown in Figure~\ref{fig:ocp_od}(b), the Mediterranean climate exhibits pronounced seasonal patterns with hot, dry summers and moderate winters, resulting in a substantially lower annual average wet-bulb temperature of \SI{11.6}{\celsius} and humidity ratio of 0.007. 
Figure~\ref{fig:ocp_pue}(b) illustrates that the Phythesis-generated design achieves an average PUE of 1.130, a 5.9\% improvement over the industry standard of 1.2 reported in Table~\ref{tab:ocp-ocp2}, while maintaining an average IT power of 6.99 MW with cooling power at 0.91 MW.
The superior PUE in Stockton compared to Jakarta directly reflects the impact of climatic conditions on cooling efficiency. The cooler, drier environment reduces the thermodynamic work required by the cooling facilities for heat rejection. 
This demonstrates that Phythesis effectively adapts design strategies to leverage local environmental advantages. Both designs achieve performance improvements over their respective industry standards while exhibiting physically consistent operational characteristics, validating the framework's capability across diverse climatic conditions for real-world industrial design purposes.

In sum, this analysis serves as an exploratory validation of Phythesis, demonstrating its capability to generate energy-efficient designs that are competitive with industry best practices under a controlled, reproducible simulation environment.
We acknowledge the inherent limitations of this comparison. OCP documentation provides only high-level settings without the detailed facility parameters, spatial configurations, or operational settings necessary for precise system modeling of the whole DCs.
Besides, our simulator cannot fully replicate actual facility conditions. 
The OCP-inspired designs represent best-effort approximations within our simulator's constraints rather than direct replicas. 
Thus, our framework shows its value as a novel framework for generative design exploration.

\onecolumn
\section{Prompt Templates}
\label{sec:prompt}
\begin{lstlisting}[
    caption={The Design LLM Prompt Template $C_\text{design}$},
    label={lst:design_prompt},
    basicstyle=\footnotesize\ttfamily,
    backgroundcolor=\color{blue!5},
    frame=single,
    breaklines=true
]
Provide a data center architecture design based on the selected facilities and equipment. You need to follow the Data Center Design Requirements while fulfilling objectives and constraints.
    
Rules: 
    - The totoal cooling load of selected ACUs must be compatible with the total cooling demand of selected racks.
    - Allocate all the selected ACUs and racks to the designed rooms.
    - Number of selected ACUs should be multiple of 2 in a room.
    - Number of selected racks should be multiple of 4 in a room.
    - A room should have at least 16 racks and 2 ACUs in total.

Unit for each attribute: 
    - powerDemand: kW
    - powerCapacity: kW
    - coolingDemand: CFM
    - coolingCapacity: CFM
    - computingCapacity: FLOPS
    - geometry: meter
    
Model Assets Library:
{assets_lib}

Weather External Inputs:
{external_inputs}

Data Center Design Requirements:
{requirements}

Previous Generated Designs and metricss:
{history}

Think step by step to select the best facility and equipment combinations and archtect the whole building strcture in <topology></topology> and <layout></layout>. If there are previous results, please try to find a better design. Prefer to select less than three types of ACUs and racks, respectively. 
<topology> and <layout> needs to fulfil the eventual design requirements.
Please conclude the final result in the following format:
<topology>
{{
    "rooms":{{
        "room_name_1":{{
            "racks": {{,
                "rack_model_1": number_of_rack_model_1,
                ...
            }},
            "acus": {{
                "acu_model_1": number_of_acu_model_1,
                ...
            }}
        }},
        ...
    }},
    "plant": {{
        "chilled_water_loop": {{
            "chillers": {{
                "chiller_model_1": number_of_chiller_model_1,
                ...
            }}
        }},
        "condenser_water_loop": {{
            "cooling_towers": {{
                "cooling_tower_model_1": number_of_cooling_tower_model_1,
                ...
            }}
        }}
    }}
}}
</topology>
<layout>
{{
    "rooms":{{
        "room_name_1": {{
            "rack_gap": room_name_1_rack_gap,
            "padding": room_name_1_padding,
            "margin": room_name_1_margin,
            "aisle_gap": room_name_1_aisle_gap,
        }},
        ...
    }}
}}
</layout>

Output:
\end{lstlisting}

\begin{lstlisting}[
    caption={The Relection LLM Prompt Template $C_\text{reflect}$},
    basicstyle=\footnotesize\ttfamily,
    backgroundcolor=\color{blue!5},
    frame=single,
    breaklines=true
]
Analyse the previous generated designs and metricss. Reflect on the design process and provide suggestions for the next design iteration. 
The summary should include the following information:
- Temperature trajectories of the data center. Is there any overheating issue?
- The power usage effectiveness (PUE) of the data center. Is the data center energy-efficient?

The suggestions should include the following information:
- New topology design.
- New spatial parameters.
- New asset selections.

Data Center Design Requirements:
{requirements}

Trajectories:
{trajectories}

The Generated Design:
{design}

Summary & Suggestions:
\end{lstlisting}

\section{Contextualized SimReady Assets}
\label{sec:asset}
\begin{lstlisting}[
    caption={The formatted physics attributes of a SimReady ACU asset.},
    basicstyle=\footnotesize\ttfamily,
    backgroundcolor=\color{blue!5},
    frame=single,
    breaklines=true
]
{
    "ACU_A": {
      "coolingCapacity": 146.8,
      "coolingType": "DX",
      "pressureRise": 460.4,
      "designWaterFlowRate": 0.003097,
      "designAirFlowRate": 13.099,
      "maximumFlowRate": 12.381,
      "designInletWaterTemperature": 16.6,
      "designOutletWaterTemperature": 23.6,
      "designInletAirTemperature": 34.7,
      "designOutletAirTemperature": 25.1,
      "designWaterTemperatureDifference": 7.9
    },
}
\end{lstlisting}
\begin{lstlisting}[
    caption={The formatted geometric attributes of a SimReady ACU asset.},
    basicstyle=\footnotesize\ttfamily,
    backgroundcolor=\color{blue!5},
    frame=single,
    breaklines=true
]
{
    "ACU_1": {
        "location": {
            "x": 0.8, 
            "y": 4.02, 
            "z": 0.0
        }, 
        "size": {
            "x": 0.995, 
            "y": 2.23, 
            "z": 0.0
        }
    }
}
\end{lstlisting}

\end{document}